\documentclass[journal]{IEEEtran}
\IEEEpubidadjcol 
\usepackage{amsmath,amsfonts}
\usepackage{algorithmic}
\usepackage{todonotes}
\usepackage{algorithm}
\usepackage{array}
\usepackage{textcomp}
\usepackage{stfloats}
\usepackage{url}
\usepackage{verbatim}
\usepackage[space, compress, sort]{cite}
\usepackage{xcolor}
\usepackage{bm}
\usepackage{microtype}
\usepackage{multirow}
\usepackage{adjustbox}
\usepackage{arydshln}
\usepackage{subcaption}
\usepackage{caption}
\usepackage{graphicx}
\usepackage{makecell}
\usepackage{float}
\usepackage{inconsolata}
\usepackage{amsmath}
\usepackage{balance} 
\usepackage{amssymb}
\usepackage{framed}
\usepackage{flushend}
\usepackage{ulem}
\definecolor{shadecolor}{rgb}{0.92,0.92,0.92}

\NewDocumentCommand{\authorbibliography}{+o+m+m+m}{%
  \IfNoValueTF{#1}{%
  }{%
    \setkeys{authorbib}{#1}%
    \immediate\write\authorbibfile{%
      \string\begin{wrapfigure}[\AuthorbibKVMacroWraplines]{\AuthorbibKVMacroImagepos}[\AuthorbibKVMacroOverhang]{\AuthorbibKVMacroImagewidth}^^J
        \string\includegraphics[scale=\AuthorbibKVMacroScale]{#2}^^J
        \string\end{wrapfigure}^^J
    }%
  }%
  \IfNoValueTF{#3}{%
    \typeout{Warning: No author name}%
  }{%
      \immediate\write\authorbibfile{%
      \unexpanded{\vspace{\AuthorbibTopSkip}}^^J
      \string\noindent\relax
      \unexpanded{\textbf{#3}\par}^^J
      \string\noindent\relax
      \unexpanded{#4}^^J%
      \unexpanded{\vspace{\AuthorbibBottomSkip}}^^J
      }%
  }%
}%

\begin{document}

\title{Robust Domain Misinformation Detection via Multi-modal Feature Alignment}
\author{Hui Liu, Wenya Wang, Hao Sun, Anderson Rocha, and Haoliang Li}

\markboth{Journal of \LaTeX\ Class Files,~Vol.~XX, No.~YY, Month~2022}%
{Shell \MakeLowercase{\textit{et al.}}: A Sample Article Using IEEEtran.cls for IEEE Journals}

\IEEEpubid{0000--0000/00\$00.00~\copyright~2021 IEEE}

\maketitle
\begin{abstract}
Social media misinformation harms individuals and societies and is potentialized by fast-growing multi-modal content (i.e., texts and images), which accounts for higher ``credibility'' than text-only news pieces. Although existing supervised misinformation detection methods have obtained acceptable performances in key setups, they may require large amounts of labeled data from various events, which can be time-consuming and tedious. In turn, directly training a model by leveraging a publicly available dataset may fail to generalize due to domain shifts between the training data (a.k.a. source domains) and the data from target domains. Most 
prior work on domain shift focuses on a single modality (e.g., text modality) and ignores the scenario where sufficient unlabeled target domain data may not be readily available in an early stage. The lack of data often happens due to the dynamic propagation trend (i.e., the number of posts related to fake news increases slowly before catching the public attention). We propose a novel robust domain and cross-modal approach (\textbf{RDCM}) for multi-modal misinformation detection. It reduces the domain shift by aligning the joint distribution of textual and visual modalities through an inter-domain alignment module and bridges the semantic gap between both modalities through a cross-modality alignment module. We also propose a framework that simultaneously considers application scenarios of domain generalization (in which the target domain data is unavailable) and domain adaptation (in which unlabeled target domain data is available). Evaluation results on two public multi-modal misinformation detection datasets (Pheme and Twitter Datasets) evince the superiority of the proposed model. The formal implementation of this paper can be found in this link\footnote{\url{https://github.com/less-and-less-bugs/RDCM}}.
\end{abstract}


\section{Introduction}
\IEEEPARstart{M}{isinformation} has become a significant concern in contemporary society, threading all aspects of individuals and society~\cite{surveyzhouxinyi, surveyfakenews}, because online social media lack serious verification processes and netizens usually cannot discriminate between fake and real news \cite{fakenewsinability}. For example, during the 2016 presidential election cycle in the United States, false news stories
claiming that Hillary Clinton ordered the murder of an FBI agent and participated in a satanic child abuse ring in a Washington pizza parlor were shared ostensibly through social media~\cite{hindman2018disinformation, willmore2016analysis}. While expert-based (e.g., PolitiFact\footnote{\url{https://www.politifact.com/}.},  GossipCop\footnote{\url{https://www.gossipcop.com/}.}) and crowd-based efforts (such as Amazon Mechanical Turk\footnote{\url{https://www.mturk.com/}.}) for manual fact-checking tools have carried precious insights for misinformation detection, they cannot scale with the volume of news on social media \cite{surveyzhouxinyi}. 

\begin{figure}[t]
\centering
\begin{subfigure}{0.235\textwidth}
\centering
\includegraphics[width=\linewidth]{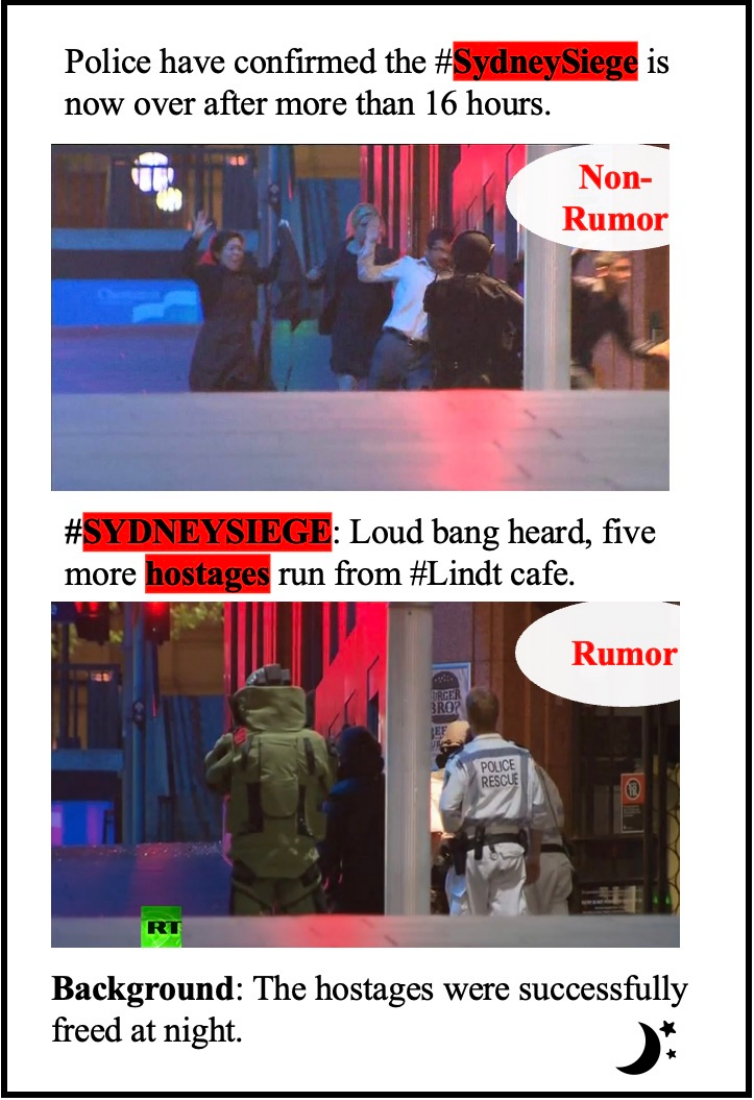}
\caption{Sydney Siege}
\label{MCA}
\end{subfigure}
\hfill
\begin{subfigure}{0.235\textwidth}
\includegraphics[width=\linewidth]{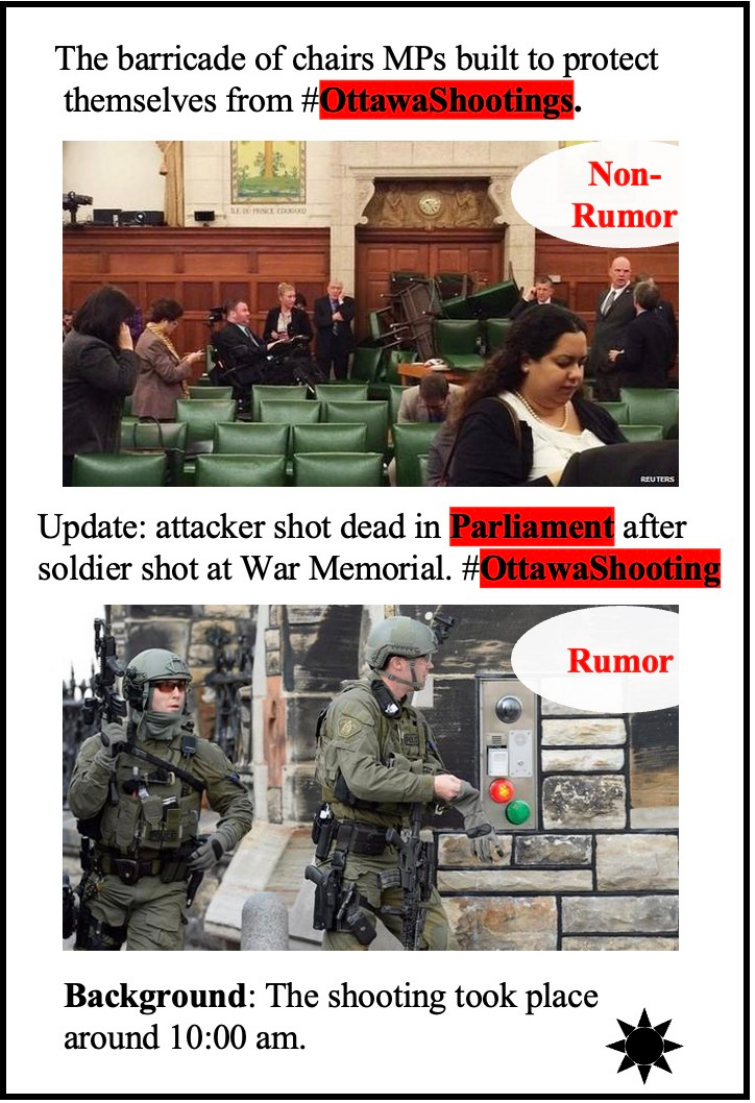} 
\centering
\caption{Ottawa Shooting}
\label{gat}

\end{subfigure}
\caption{Examples of Sydney Siege and Ottawa Shooting domains from Pheme Dataset. Sydney Siege was a terrorist attack in which a gunman held hostage ten customers and eight employees in Sydney on December 15-16, 2014. Ottawa Shooting took place on Ottawa's Parliament Hill, leading to the death of a Canadian soldier on October 22, 2014.}
\label{exampleofdomains}
\end{figure}

\IEEEpubidadjcol
Various methods have been proposed to perform misinformation detection based on textual features \cite{textualfeature1, textualfeature2,gaojingaaai20} and propagation patterns \cite{network1, network2,network3}. As the increasing misinformation with images disseminates more quickly and is more believable, another line of exploration \cite{multi-modal1, multi-modal2, multi-modal3,multi-modal6} exploits multi-modal features to verify misinformation. Despite the success of these algorithms, they typically require considerably large labeled datasets, which may not be feasible for real-world applications as data collection and annotation can be cumbersome and time-consuming. 

Moreover, directly training with large-scale datasets may not generalize well to unseen events on account of the domain shift~\cite{gaojingkdd21, tkde2022, drkdd2018eann, daxsctmm2020}, as there exist discrepancies between data distributions across different domains, such as word frequency and image style as Fig.~\ref{exampleofdomains} depicts. For example, ``Sydney Siege" and ``hostages" frequently occur in the Sydney Siege event\footnote{\url{https://en.wikipedia.org/wiki/Lindt_Cafe_siege.}}, while ``Parliament" and ``Ottawa" for Ottawa Shooting\footnote{\url{https://en.wikipedia.org/wiki/2014_shootings_at_Parliament_Hill,_Ottaw.}}. Additionally, the illumination conditions are dark and bright for these two events, caused by the different times of occurrence.

Recent studies resort to transfer learning to learn domain-robust misinformation detector through mitigating the distribution discrepancy between the source (a.k.a., training data) and the target domain (a.k.a., testing data)~\cite{draaai21aimila,drkdd2018eann}. 
However, there still exist two main limitations. First, intuitively, during the dissemination of a specific news event, the number of relevant posts increases slowly at first and rapidly when catching significant public attention~\cite{nature1, nature2}. This indicates we cannot obtain sufficient data for the target domain early on. Hence, the methods above cannot be swiftly applied in this case as they require labeled~\cite{daICBD2021shukai, daskwww2022, daxsctmm2020} and unlabeled target domain data~\cite{daICBD2021shukai, daskwww2022, daxsctmm2020, draaai21aimila,drkdd2018eann,gaojingkdd21, tkde2022} to be available during training. Secondly, existing methods for cross-domain misinformation detection ignore the issue of discrepancy between visual and textual modalities. We argue that directly performing distribution alignment across domains without considering the gap between different modalities may not be optimal for capturing robust domain information for multi-modal misinformation detection.  

We propose a unified, robust domain and cross-modality framework named \textbf{RDCM} for multi-modal misinformation detection that seeks to address the limitations above. The unified framework can be applied to two application scenarios: 1) real-time misinformation detection {(i.e., when target domain data are not accessible during training, corresponding to domain generalization)}; and 2) offline misinformation detection (i.e., when unlabeled target domain data are available during training, which corresponds to domain adaptation). 

To align multi-modal distributions and mitigate the modality gap between source and target domains, we propose to leverage an inter-domain alignment module based on the joint distribution of textual and visual features and a cross-modality alignment module based on contrastive learning for the multi-modal misinformation detection task. The inter-domain alignment module measures the joint distribution of modalities (i.e., image and text) based on the kernel mean embedding, reproducing the kernel Hilbert space (RKHS)~\cite{mmd1} and then aligns the joint distribution of different domains by minimizing the corresponding Maximum Mean Discrepancy (MMD)~\cite{mmd2}.  

We align distributions among multiple source domains for the scenario which requires real-time applications (a.k.a. domain generalization) and align distributions between each source and the target domain for the scenario where misinformation detection can be performed offline (a.k.a. domain adaptation).

Inspired by contrastive learning in self-supervised tasks~\cite{mocov1, simclr1, clip}, the cross-modality alignment module exploits contrastive learning to bridge the modality gap with a novel sampling strategy tailored for multi-modal misinformation detection. After inter-domain and cross-modal (i.e., feature alignment across different modalities in a single domain) alignment, we expect to extract domain-invariant textual and visual features of multi-modal posts and concatenate them for misinformation detection. The empirical study shows that our model yields state-of-the-art results on two public datasets.

The key contributions of this work are:
\begin{itemize}
\item A unified framework that tackles the domain generalization (target domain data is unavailable) and domain adaptation tasks (target domain data is available). This is necessary as obtaining sufficient unlabeled data in the target domain at an early stage of misinformation dissemination is difficult;

\item Inter-domain and cross-modality alignment modules that reduce the domain shift and the modality gap. These modules aim at learning rich features that allow misinformation detection. Both modules are plug-and-play and have the potential to be applied to other multi-modal tasks.

\end{itemize}

\section{Related Work}
This section reviews domain generalization (DG), domain adaptation (DA), and robust domain misinformation detection.

\subsection{Domain Generalization and Domain Adaptation}
Supervised machine learning algorithms assume similar training and testing distributions, but practical deployment requires models to generalize well on unseen, out-of-distribution data. Domain generalization (DG) and domain adaptation (DA) address this challenge. DG learns from one or multiple source domains, while DA requires access to target domain data during training, making DG more difficult.

Domain generalization is widely used in computer vision and natural language processing. A recent survey~\cite{wang2022generalizing} classified DG methods into three categories: data manipulation, representation learning, and learning strategy. 

Data manipulation involves generating samples through data augmentation~\cite{dataagu1, dataagu2} or data generation methods~\cite{dggneration1} to increase the diversity and quantity of source domain data.

Representation learning works are inspired by the theory that domain invariant representations are transferable to unseen domains \cite{theory1}. These works aim to learn robust domain representation extraction functions by either aligning feature distributions among source domains \cite{dghaoliang1,dgrel1,dgrel2} or disentangling features into different sub-spaces (domain-specific and domain-sharing space) \cite{dgrel3, dgrel4}. For instance, Li et al. \cite{dghaoliang1} used adversarial autoencoders with Maximum Mean Discrepancy (MMD) distance to align distributions across different domains and learn a generalized latent feature representation. Ding and Fu \cite{dgrel3} designed domain-specific and domain-sharing networks for the disentanglement in individual domains and across all domains, respectively.

Finally, the learning strategy-based DG methods focus on machine learning paradigms to enhance the generalization performance, such as meta-learning \cite{mldg}, ensemble learning \cite{dgrel5}, gradient-based DG \cite{fish}, among others.  

Domain adaptation methods differ from domain generalization in that they require access to target domain data during the training process \cite{coral,DAN,DANN}. These methods are categorized into two groups for single source domain visible during adaptation (SDA). One group uses explicit discrepancy measures, like H-divergence \cite{ben2010theory}, MMD \cite{mmd1, mmd2}, Wasserstein Distance \cite{wasserstein1,wasserstein2}, and second-order statistics \cite{coral}, to reduce the shift between source and target distributions. The other group employs adversarial learning, where a domain discriminator is confused in a min-max manner \cite{dahaoliang2}, to implicitly align the source and target distributions. Additionally, early theoretical analysis \cite{msatheory1, msatheory2} demonstrated that minimizing a weighted combination of source risks can achieve lower target error.

The above methods can also be applied when data from multiple source domains are available during training (MDA). Peng et al. \cite{msa1cvpr19} dynamically aligned moments of feature distributions of multiple source domains and the target domain with theoretical insights. Zhu et al. \cite{msa2aaai19} proposed a two-stage alignment framework that aligned distributions of each pair of source
and target domains and the outputs of classifiers. Despite the progress, effectively applying DG and DA methods to multi-modal settings with large semantic gaps among different modalities remains unsolved. 

\subsection{Robust Domain Misinformation Detection}
The widespread presence of misinformation on social media has escalated the issue of social distrust, drawing significant attention from both society and the research community. However, many existing misinformation detection methods \cite{textualfeature1, textualfeature2, network1, network2, network3, multi-modal1, multi-modal2, multi-modal3, tifsfakenews} are domain-specific and may not perform effectively on unseen domains due to the domain shift. Moreover, these methods require extensive and diverse training data, which is impractical given the rapid accumulation of events and news.

Robust domain methods were developed aiming at the domain shift. Some work \cite{daskwww2022, daxsctmm2020, daICBD2021shukai, draaai21aimila} fall into domain adaptation, assuming access to target domain data during training. For instance, Mosallanezhad et al. \cite{daskwww2022} proposed a domain adaptive detection framework using reinforcement learning and incorporating auxiliary information. Silva et al. \cite{draaai21aimila} introduced an unsupervised technique for selecting unlabeled news records to maximize domain coverage and preserve domain-specific and cross-domain knowledge through disentangle learning.

However, these methods may not accommodate the dynamic nature of misinformation generation and propagation, where target domain data might be unavailable during training. Limited access to timely target domain data hinders their real-time application. Another group of works explores using powerful search engines (e.g., Google) to retrieve background knowledge for fact-checking \cite{openset1, openset2}. Yet, unverified online information introduces noise that can negatively impact performance.

\begin{figure*}[t]
\centering
\includegraphics[width=0.8\linewidth]{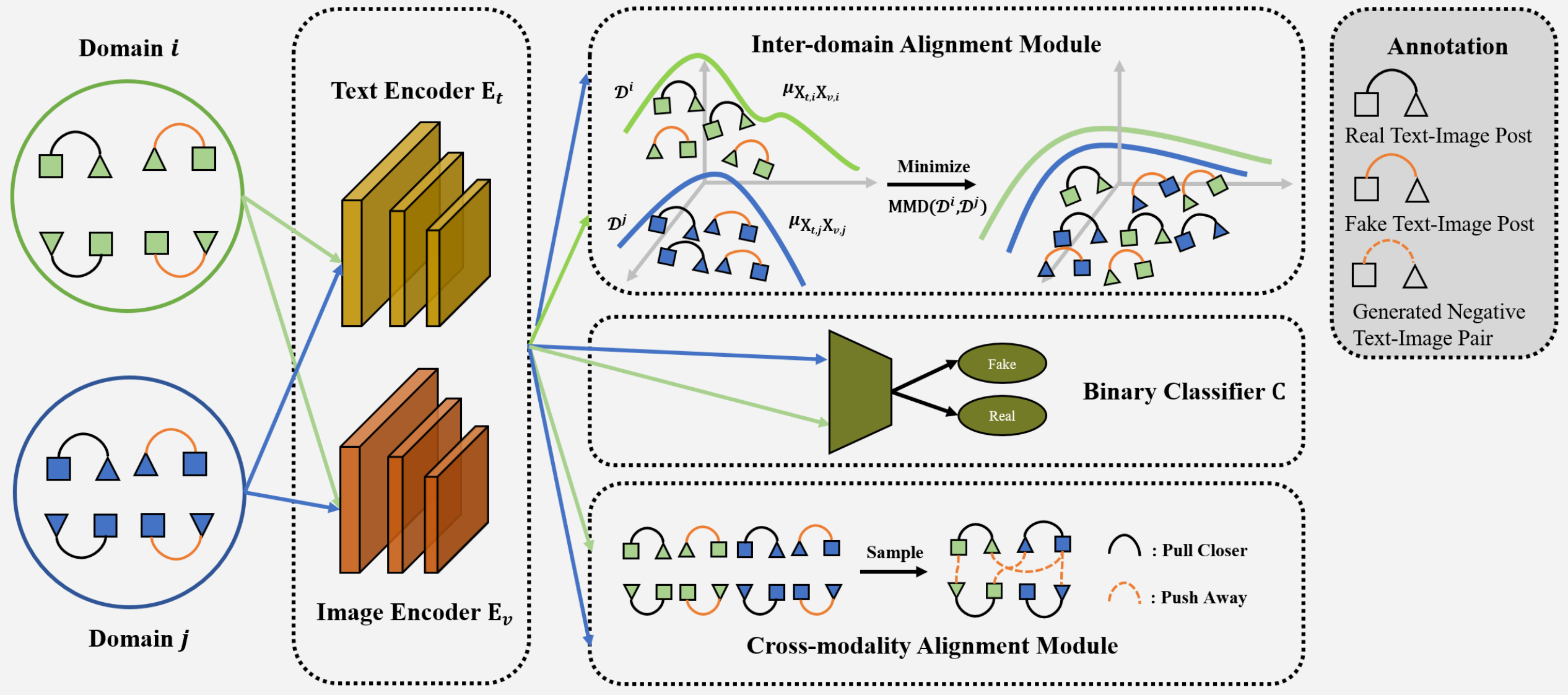}
\label{stj}
\caption{Proposed robust domain and cross-modal framework. In the DG setup, we take multiple source domains as input and extract textual and visual features through the Text Encoder and the Image Encoder. Then we align the joint distributions of textual and visual modalities between each source domain pair by Inter-domain Alignment Module, reduce the modality gap by Cross-modality Alignment Module,  and detect misinformation of source domains through Binary Classification. The DA setup takes multiple sources and the target domain as input. Compared with DG, it further aligns joint distributions between each source domain and the target domain but only performs cross-modal alignment and trains the classifier on source domains.}
\label{overview}
\end{figure*}

\section{Proposed method}
The multi-modal multi-domain misinformation detection framework comprises four components: Multi-modal Representation Extraction (Text and Image Encoders), Inter-domain Alignment, Cross-modality Alignment, and Classification. Textual and image features are extracted from a post using the corresponding encoders. The Inter-domain Alignment module removes domain-specific information while preserving domain-agnostic information. The Cross-modality Alignment combines textual and visual representations. The combined domain-robust and modality-aligned features are then used for misinformation detection. While designed for domain generalization (DG), the framework can be extended to unsupervised domain adaptation (DA) by adapting the inter-domain module to align distributions between source and target domains.

\subsection{Task Definition}
\IEEEpubidadjcol
The goal of multi-modal misinformation detection is to determine the authenticity of a text and an associated image, classifying the pair as fake (rumor) or real (non-rumor). To address challenges posed by fast-emerging events and costly annotations, researchers have explored various domain adaptation methods~\cite{daxsctmm2020, daskwww2022, daICBD2021shukai, daICBD2021shukai, draaai21aimila} to learn robust domain features and mitigate domain shifts. However, these methods overlook the difficulty of collecting sufficient data in the target domain during the early stages of fake news dissemination and fail to consider the presence of multiple modalities in real-world news pieces. To address these issues, we propose a unified framework to handle the multi-modal misinformation detection task, making it suitable for both domain generalization (DG) and domain adaptation (DA) scenarios.

Formally, given $\mathcal{D}_S= \left\{ \mathcal{D}^1_S, \mathcal{D}^2_S, \ldots, \mathcal{D}^M_S \right\}$ the collection of M labeled source domains and $\mathcal{D}_T$ the unlabeled target domain where all domains are defined based on different news events, our method aims to find a hypothesis in the given hypothesis space, which minimizes the classification error on $\mathcal{D}_T$. Each source domain can be represented as $\mathcal{D}_S^m = \left\{ {(t^m_n, v^m_n), y^m_n} \right\}_{n=1}^{N_m}$ and the target domain can be denoted as $\mathcal{D}_T = \left\{ \left(t_n, v_n \right) \right\}_{n=1}^{N_T}$, where $N_m$ ($1 \leq m \leq M$) is the number of samples in the $m$-th source domain, $N_T$ is the number of samples in the target domain, and $y \in \{0, 1 \}$ is the gold label (1 indicates fake information for the Twitter Dataset or the rumor for the Pheme Dataset and 0 otherwise). Additionally, $\left(t, v\right)$ is a text-image pair, where $t$ is a text sentence, and $v$ is the corresponding image. We assume \textbf{no availability} of target domain data $\mathcal{D}_T$ in the scenario of DG. 

\subsection{Multi-modal Representation Extraction}
Given an input text-image pair $(t, v)$\footnote{We omit the subscript for simplicity unless specifically stated.} in each domain, following previous work \cite{drkdd2018eann, gaojingkdd21}, we leverage a convolutional neural network (i.e., TextCNN \cite{textcnn}) with an additional two-layer perceptron (MLP) as the textual encoder to obtain the representation of $t$ as $\mathbf{x}_t$:
\begin{equation}
\label{text-feature}
    \mathbf{x}_t = \textbf{E}_t(t; \bm{\theta}_{t} ),
\end{equation}
where $\mathbf{x}_t \in \mathbb{R}^d$ is the final representation of $t$, $\textbf{E}_t$ represents the textual encoder, and $\bm{\theta}_{t}$ represents the parameter of TextCNN and corresponding MLP. As large-scale pre-trained models have excelled in natural language processing tasks, we adopt the word embedding extracted by RoBERTa~\cite{roberta} as initializing word vectors of TextCNN, following existing work~\cite{nan2022improving, gaojingkdd21 }. The reason why we do not fine-tune RoBERTa is to avoid over-parameterization, which may harm the generalization ability of the model.

For image representation, given an image $v$, following existing methods \cite{daICBD2021shukai,daxsctmm2020,gaojingaaai20}, we use ResNet50 as the visual backbone neural network and choose the feature of the final pooling layer as the initial visual embedding. Then, similar to the text modality, we also use a MLP to reduce its dimension to $d$ given as
\begin{equation}
\label{image-feature}
    \mathbf{x}_v =  \textbf{E}_v(v; \bm{\theta}_{v} ),
\end{equation}
where $\mathbf{x}_v \in \mathbb{R}^d$ is the final representation of the image $v$, $\textbf{E}_v$ represents the visual encoder, and $\bm{\theta}_{v}$ represents the parameter of ResNet50 and the visual MLP. 

We use $\textbf{X}_t$ and $\textbf{X}_v$ to denote random variables instantiated by $ \mathbf{x}_t$  and by $ \mathbf{x}_v$ in one domain. After extracting the textual and visual features of each text-image pair for multiple source domains $\{\mathcal{D}^m_S\}_{1\leq m \leq M}$ and target domain $\mathcal{D}_T$, we can empirically estimate the probability distribution of textual features $\mathbb{P}(\mathbf{X}_t)$ and the probability distribution of visual features  $\mathbb{P}(\mathbf{X}_v)$ by drawing samples i.i.d. from variables $\mathbf{X}_t$ and $\mathbf{X}_v$ from each domain. 

\subsection{Multi-modal Feature Alignment}

Multi-modal feature alignment aims to extract robust domain information for misinformation detection; as such, the trained model can be better generalized to unseen events. However, existing cross-domain-based methods for misinformation detection can be limited as most of them only focus on a single modality for misinformation detection. While one can perform marginal distribution alignment on textual features $\mathbf{X}_t$  and visual features $\mathbf{X}_v$, separately, or perform distribution alignment through feature concatenation or element-wise production~\cite {daxsctmm2020, daskwww2022, yang2022multi-modal}, the correlation property across multiple modalities has been ignored, which may hinder robust domain misinformation detection when having textual and visual information as input. To tackle this limitation, we propose to explore domain covariance information on both the event level (i.e., domain) and sample level, corresponding to Inter-domain Alignment and  Cross-modality Alignment, respectively.

\subsubsection{Inter-domain Alignment}
Among various inter-domain alignment methods based on distribution measurement, Maximum Mean Discrepancy (MMD) \cite{mmd2} has been proven to be effective where the distribution of samples can be formulated through kernel mean embedding \cite{mmd1} in a non-parametric manner.  One intuitive way is to align the marginal distribution of textual and visual modality across domains through MMD, which can be defined as 
\begin{equation}
\small
    \textrm{MMD}(\mathcal{D}_S^i,\mathcal{D}_S^j) = \|\mu_{\mathbf{X}_{t,i}}-\mu_{\mathbf{X}_{t,j}}\|_\mathcal{H}^2 + \|\mu_{\mathbf{X}_{v,i}}-\mu_{\mathbf{X}_{v,j}}\|_\mathcal{H}^2.
\end{equation}
\vspace{-2pt}
We use samples from the $i$-th and $j$-th source domains as example. $\mu$ denotes the kernel mean embedding operation in reproducing kernel Hilbert space (RKHS) $\mathcal{H}$ \cite{mmd1}, which is to compute the mean of latent features in the RKHS as $\mu_{\mathbf{X}}(\mathbb{P}):= \textrm{E}_\mathbf{X}[\phi(\mathbf{X})]  = \int_{X} \phi(x)\, {\rm d}\mathbb{P}(x)$ and $\phi$ denotes a kernel function. Here $\mu_{\mathbf{X}_{t,i}}$ and $\mu_{\mathbf{X}_{v,j}}$ indicate the textual mean embedding for the $i$-th source domain and the visual mean embedding for the $j$-th source domain, respectively.  However, directly performing marginal distribution alignment may not capture the correlation information between textual and visual modalities. We propose to align the joint feature distribution upon textual and visual modalities where the kernel mean embedding can be formulated through the covariance operator $\otimes$ on RKHS \cite{jointdis} as 
\begin{equation}
\label{jkernelemb}
     \mu_{_{\mathbf{X}_t,\mathbf{X}_v}} = \textrm{E}[ \phi_t(\mathbf{X}_t) \otimes \phi_v(\mathbf{X}_v)]. 
\end{equation}
We can better capture the cross-covariance dependency between textual and visual modalities, contributing to robust domain multi-modal misinformation detection, and the new inter-domain alignment MMD can be formulated as
\begin{equation}
\begin{aligned}
    \textrm{MMD}(\mathcal{D}_S^i,\mathcal{D}_S^j) = \|\mu_{\mathbf{X}_{t,i},\mathbf{X}_{v,i}}-\mu_{\mathbf{X}_{t,j},\mathbf{X}_{v,j}}\|_\mathcal{H}^2.
\end{aligned}
\end{equation}
We seek for the empirical estimate of $\textrm{MMD}(\mathcal{D}_S^i,\mathcal{D}_S^j)$ \cite{jointdis} which can be computed as
\begin{equation}
\small
\begin{aligned}
\textrm{MMD}(\mathcal{D}_S^i,\mathcal{D}_S^j)= \frac{1}{N_{i}^2} \sum_{p=1}^{N_{i}}\sum_{q=1}^{N_{i}}  k_v(\mathbf{x}_{v,i,p}, \mathbf{x}_{v,i,q})k_t(\mathbf{x}_{t,i,p}, \mathbf{x}_{t,i,q})  \\
+ \frac{1}{N_j^2} \sum_{p=1}^{N_j}\sum_{q=1}^{N_j}  k_v(\mathbf{x}_{v,j,p}, \mathbf{x}_{v,j,q})k_t(\mathbf{x}_{t,j,p}, \mathbf{x}_{t,j,q}) \\
- \frac{2}{N_i N_j} \sum_{p=1}^{N_i}\sum_{q=1}^{N_j}  k_v(\mathbf{x}_{v,i,p}, \mathbf{x}_{v,j,q})k_t(\mathbf{x}_{t,i,p}, \mathbf{x}_{t,j,q}), 
\end{aligned}
\end{equation}
where $\mathbf{x}_{v,i,p}$ denotes the latent feature of the $p$-th sample from the modality $v$ of the domain $i$, $k_t$ and $k_v$ are Gaussian kernel functions to map extracted features $\mathbf{x}_{t}$ and $\mathbf{x}_{v}$ into RKHS, corresponding to the textual modality and visual modality, respectively. 

Assume we have training samples from $M$ different domains as $\mathcal{D}_S= \left\{ \mathcal{D}^1_S, \mathcal{D}^2_S, \ldots, \mathcal{D}^M_S \right\}$, the inter-domain alignment loss based on data collected from textual and visual modalities can be formulated as
\begin{equation}\label{inter1}
\small
    \mathcal{L}_{\textrm{inter}} = \binom{2}{M} \sum_{i=1}^{M-1}\sum_{j=i+1}^M \textrm{MMD}(\mathcal{D}^i_S, \mathcal{D}^j_S).
\end{equation}
When some data for testing are available during training, we can extend the inter-domain alignment loss above  by incorporating it with target domain data $D^T$ as
\begin{equation}
\small
\begin{aligned}
\label{inter2}
\mathcal{L}_{\textrm{inter}} &= \binom{2}{M} \sum_{i=1}^{M-1}\sum_{j=i+1}^M \textrm{MMD}(\mathcal{D}^i_S, \mathcal{D}^j_S) \\&+  \binom{1}{M} \sum_{i=1}^{M} \textrm{MMD}(\mathcal{D}^i_S, \mathcal{D}_T).
\end{aligned}
\end{equation}

\subsubsection{Cross-modality Alignment}
Besides exploring domain-wise correlations between the textual and visual modalities, we are also interested in mining sample-wise correlations (i.e., aligning the textual and visual modalities based on a single sample). Hence, we propose a novel contrastive loss for the cross-modality alignment module to model pairwise relations between texts and images by pulling semantically similar pairs closer while pushing dissimilar ones away. Though recent vision-language contrastive learning methods have shown promising results in learning meaningful representations~\cite{clip, tosci}, their sampling strategies for drawing positive and negative pairs may not be suitable for misinformation detection. Specifically, existing sampling methods derive positive pairs from the original input and negative pairs via random sampling in one minibatch. However, in our setting, cross-modal correspondence or similarity is more likely to only exist in real news rather than in misinformation scenarios. Besides, texts for different misinformation examples may use the same image in a specific event, which results in the image and text of many negative samples being close to each other in the semantic space. The observations above motivate us to design a metric for text-image similarity measurement, which can be further utilized for negative sample selection, contributing to cross-modality alignment through contrastive learning. 

To tackle the above problems, we propose a novel sampling strategy by only taking real posts as positive samples and filtering out negative samples with high semantic similarity on the visual modality with a weighting function as follows:
\begin{equation}
\footnotesize
\label{indicator}
\mathbb{I}((\textbf{x}_{t,p}, \textbf{x}_{v,p}),(\textbf{x}_{t,q}, \textbf{x}_{v,q})) = \begin{cases} 
0,  & \text{if } \textrm{sim}(\mathbf{h}_p, \mathbf{h}_q) \ge \beta \\
\beta - \textrm{sim}(\mathbf{h}_p, \mathbf{h}_q) & \textrm{else}, 
\end{cases}
\end{equation}

Here $p$ and $q$ denote indices corresponding to  the $p$-th and $q$-th samples in a minibatch, $\mathbf{h}_p$ and $\mathbf{h}_q$ denote the output of feature processing on $\mathbf{x}_{v,p}$ and  $\mathbf{x}_{v,q}$ respectively\footnote{We omit the domain index here for simplicity.}.  $\text{sim}(\mathbf{h}_p, \mathbf{h}_q) =( \frac{\mathbf{h}_p \mathbf{h}_q^\top}{\lVert \mathbf{h}_p \rVert  \lVert \mathbf{h}_q \rVert}  + 1) /2$ represents the similarity between $(\textbf{x}_{t,p}, \textbf{x}_{v,p})$ and $(\textbf{x}_{t,q}, \textbf{x}_{v,q})$, and $\beta$ is a threshold to remain semantic dissimilar pairs as negative samples. Regarding the feature processing function, one can choose an identity mapping for feature processing on visual modality. However, for the problem of misinformation detection, we are more interested in the instance-level information (i.e., object) instead of semantic information contained in the latent features. As a result, we take for $\mathbf{h}$ the output of the softmax layer of the backbone for the visual modality (e.g., ResNet50 in our model) which can measure the instance-level similarity between images well \cite{instancecon}.

Especially, it is a good surrogate for similarity between $\mathbf{x}_{t,p}$ and $\mathbf{x}_{v,q}$ when we assume  $\mathbf{x}_{t,p}$ and $\mathbf{x}_{v,p}$  of real posts are semantically relevant.

After performing a sample section to get the positive and negative text-image pairs, we leverage the contrastive loss objective in \cite{simclr1} and enhance it by our weighting function to learn cross-modal semantic alignment on source domains $\mathcal{D}_S$, which can be formulated as follows:
\begin{equation}
\label{intra}
\footnotesize
\mathcal{L}_{intra} = - \textrm{log} \frac{e^{\frac{\tilde{\mathbf{x}}_{t,p} \tilde{\mathbf{x}}_{v,p}^\top}{\tau}}}{e^{\frac{\tilde{\mathbf{x}}_{t,p} \tilde{\mathbf{x}}_{v,p}^\top}{\tau}} + \sum_{q\ne p}^B e^{ \frac{\tilde {\mathbf{x}}_{t,p} \tilde{\mathbf{x}}_{v,q}^\top}{\tau}} \mathbb{I}((\textbf{x}_{t,p}, \textbf{x}_{v,p}),(\textbf{x}_{t,q}, \textbf{x}_{v,q}))},
\end{equation}
where $p$ represents the indices of real posts in a minibatch, $q$ represents the indices of the other samples in this minibatch except the $p$-th sample, $B$ is the minibatch size, and $\tau$ is a temperature hyperparameter. Additionally, we normalize $\mathbf{x}_t$ and $\mathbf{x}_v$ to $\tilde{\mathbf{x}}_{t}$ and $\tilde{\mathbf{x}}_{v}$ based on $L_2$ normalization to restrict the range of similarity scores, which have been widely adopted in \cite{simclr1, mocov1, conbaseline1}. Compared with the original loss in \cite{simclr1}, our proposed $\mathcal{L}_{intra}$ can further push  $\mathbf{x}_{v,q}$ of the hard negative samples far away from corresponding $\mathbf{x}_{t,p}$ in the shared feature space and mitigate the influence of inappropriate random sampling for multi-modal tasks \cite{conbaseline1, multi-modal2} to perform better modality alignment. 

\subsection{Classification}
Given the textual feature $\mathbf{x}_{t}$ and visual feature  $\mathbf{x}_{v}$ of one post $(t,v)$ in source domains $\mathcal{D}_S$, we concatenate them for the final prediction:
\begin{equation}
\hat{\mathbf{y}} = \textbf{C}(\mathbf{x}_{t}, \mathbf{x}_{v}; \bm{\theta}_c).     
\end{equation}

Here \textbf{C} is a classifier consisting of a MLP followed by a softmax activation function,  $\bm{\theta}_c$ is its parameters, and $\hat{\mathbf{y}}$ is the predicted label. Then, the classifier is trained with cross-entropy loss against the ground-truth label $\mathbf{y}$ on source domains $\mathcal{D}_S$ as $\mathcal{L}_{cls} = -\mathbf{y}\textrm{log}(\hat{\mathbf{y}})$, in which label 1 represents fake posts (rumors) while 0 means real posts (non-rumors) in our task.

In this work, we are especially concerned with the robust domain multi-modal misinformation detection that requires a model to simultaneously map textual and visual features into domain-invariant and modality-aligned semantic space to improve classification performance. As such, we combine $\mathcal{L}_{inter}$ in Eq.~\ref{inter1}, $\mathcal{L}_{intra}$  in Eq.~\ref{intra} and $\mathcal{L}_{cls}$ as the final form of our training objective in DG situation:
\begin{equation}
\label{loos1}
    \mathcal{L} = \lambda_1 \mathcal{L}_{inter} + \lambda_2 \mathcal{L}_{intra} + \mathcal{L}_{cls},
\end{equation}
where $\lambda_1$ and $\lambda_2$ are weighting parameters to balance the importance of $\mathcal{L}_{inter}$, $\mathcal{L}_{intra}$ and $\mathcal{L}_{cls}$. Moreover, we can easily extend our method to the DA situation by replacing $\mathcal{L}_{inter}$ in Eq.~\ref{inter1} as $\mathcal{L}_{inter}$ in Eq.~\ref{inter2} without changing our framework.

\section{Experiments}
We devise experiments to answer the following research questions. For conciseness, \textbf{RQ4} and \textbf{RQ5}, the elaborate analysis of the inter-domain alignment module and cross-modality alignment module, are explained in the Appendix.
\begin{itemize}
\item \textbf{RQ1:} Do unlabeled target domain data and multiple modalities boost domain misinformation detection?

\item \textbf{RQ2}: How effective is the proposed robust domain and cross-modal detection method (\textbf{RDCM}) compared with existing methods for misinformation detection? 

\item \textbf{RQ3:} How do the components of \textbf{RDCM} affect results? 

\item \textbf{RQ4:} How effective is the method to mitigate the domain shift by aligning the joint distribution of text and visual features represented by kernel mean embedding?

\item \textbf{RQ5:} How effective is the sampling strategy for the cross-modality alignment module? 
\end{itemize}

\subsection{Data Preparation}
We adopt two benchmark datasets, Pheme~\cite{pheme} and Twitter~\cite{Twitter}, to validate the effectiveness of the proposed misinformation detection approach~\textbf{RDCM}. 

Pheme dataset is constructed by collecting tweets related to five breaking news events: \textit{Charlie Hebdo}, \textit{Sydney Siege}, \textit{Ferguson Unrest}, and \textit{Ottawa Shooting} and \textit{Germanwings Crash}. 
As the original Pheme dataset does not include images, we obtain relevant images through the Twitter API using the tweet ID contained in each sample if the sample has attached images, following~\cite{daxsctmm2020}. In this work, we detect misinformation by incorporating text and image information. Thus, we remove the tweets without any text or image and finally get four event domains. If multiple images are attached to one post, we randomly retain one image
and discard the others. The detailed statistics are listed in Table~\ref{datasetpheme}.

The Twitter dataset collects text content, attached images/videos, and social context information related to 11 events. However, several events are removed from the experiments because of only having real or fake posts. Following the data cleaning method for the Pheme dataset, we only preserve samples containing texts and images and obtain four event domains, including \textit{Hurricane Sandy}, the \textit{Boston Marathon bombing}, \textit{Malaysia}, and \textit{Sochi Olympics}. It is worth noting that many samples have the same image in this dataset, which challenges the generation of negative multi-modal pairs for contrastive learning. The detailed statistics are listed in Table \ref{datasettwitter}.  

Regarding the criterion of labels, in the Pheme dataset, the sample is labeled as a rumor when it is
unverified\footnote{One post is defined as unverified when there is no evidence supporting it (e.g., logically self-consistent between the text and image) or there is no official confirmation from authoritative sources.} at the time of posting, it is labeled as non-rumor when it belongs to 
the other circulating information \cite{phemeset, phemeset_2}. Moreover, in the Twitter dataset, the sample is identified as fake when it shares an image that does not represent the event it refers to (e.g., maliciously tampering with images and reposting previously captured images in a different event). At the same time, they are considered real when it shares an image that legitimately represents the event it refers to  \cite{Twitter}. As a result, a huge discrepancy exists between domains from different datasets.

To further verify the generalization of the proposed approach, we conduct Cross-dataset experiments between these two datasets. Especially we select three source domains from either
the Pheme or Twitter dataset to train the model and evaluate its performance on the target domain
from the other dataset. Finally, the results of four cases $\mathcal{COF}\rightarrow \mathcal{M}$, $\mathcal{CSF}\rightarrow \mathcal{A}$, $\mathcal{ABI}\rightarrow \mathcal{S}$ and $\mathcal{ABI}\rightarrow \mathcal{O}$ are reported in our experiments.
\begin{table}[H]
  \caption{Statistics of Pheme Dataset}
  \centering
  \label{datasetpheme}
  \begin{adjustbox}{max width=1.0\columnwidth}
  \begin{tabular}{c|ccc}
    \hline
    Event&Rumor&Non-Rumor&All\\
    \hline
    Charlie Hebdo ($\mathcal{C}$)&181&742&923\\
    Sydney Siege ($\mathcal{S}$)&191&228&419\\
    Ferguson Unrest ($\mathcal{F}$)&42&309&351\\
    Ottawa Shooting ($\mathcal{O}$)&146&110&256\\
    \hline
\end{tabular}
\end{adjustbox}
\end{table}

\begin{table}[H]
  \caption{Statistics of Twitter Dataset}
  \centering
  \label{datasettwitter}
  \begin{adjustbox}{max width=1.0\columnwidth}
  \begin{tabular}{c|ccc}
    \hline
    Event&Fake&Real&All\\
    \hline
    Hurricane Sandy ($\mathcal{A}$)&5461&6841&12302\\
    Boston Marathon bombing  ($\mathcal{B}$)&81&325&406\\
    Malaysia ($\mathcal{M}$)&310&191&501\\
    Sochi Olympics  ($\mathcal{I}$)&274&127&398\\
    \hline
\end{tabular}
\end{adjustbox}
\end{table}

\subsection{Experimental Setup}
\subsubsection{Baselines}
For comparison purposes, we adopt baselines from four categories: uni-modality, multi-modality, domain generalization, and domain adaptation baselines. 

Uni-modality baselines comprise \textbf{TextCNN-rand}, \textbf{TextCNN-roberta}, \textbf{Bert} \cite{bert}, 
and \textbf{ResNet} \cite{resnet}. \textbf{TextCNN-rand}, \textbf{TextCNN-roberta}, and \textbf{Bert} are text modality-based models which only exploit textual information for classification. Both \textbf{TextCNN-rand} and \textbf{TextCNN-roberta} are based on TextCNN framework \cite{textcnn}. Their difference is that the workpiece embedding of \textbf{TextCNN-rand} uses random initialization, and \textbf{TextCNN-roberta} is initialized from the RoBERTa-base\footnote{\url{https://huggingface.co/roberta-base}}, which is frozen during training, following \cite{drkdd2018eann, daICBD2021shukai}. \textbf{Bert} is a  transformer-based pre-trained model, and we utilize one of its variants\footnote{\url{https://huggingface.co/bert-base-uncased}} to generate the embedding of [CLS] token for detection. We compare the model with the visual modality method~\textbf{ResNet}~\cite{resnet}, which replaces the final classification layer as a binary classification layer for misinformation detection.

Multi-modality baselines include \textbf{Vanilla}~\cite{vanilla} and \textbf{ModalityGat}~\cite{multimodalgat}, which take TextCNN and ResNet as textual and visual encoders, respectively. \textbf{Vanilla} concatenates textual and visual features to perform classification, similar to our proposed method without Inter-domain Alignment and  Cross-modality Alignment components. On the other hand, \textbf{ModalityGat} introduces a gate mechanism to fuse the information from different modalities based on their corresponding importance. 

Domain generalization baselines consist of  \textbf{EANN} \cite{drkdd2018eann}, \textbf{IRM} \cite{irm}, \textbf{MLDG}\cite{mldg} and \textbf{Fish}\cite{fish}, among which the first two belong to representation learning based DG, and the last two belong to learning strategy based DG. In detail, \textbf{EANN} confuses an event domain discriminator in an adversarial manner to learn shared features among multiple events. \textbf{IRM} aims to estimate invariant and causal predictors from multiple source domains to improve the generalization performance on the target domain. \textbf{MLDG} is a meta-learning framework that simulates domain shift by synthesizing virtual meta-train and meta-test sets in each mini-batch. Finally, \textbf{Fish} matches the distribution of many source domains by maximizing the inner product between gradients of these domains. While there exists some work using data augmentation to improve the robustness of misinformation detection based on social networks \cite{drwww2019majinggal, luo2021newsclippings}, these works are not designed for multi-modal based misinformation detection, and how to perform suitable data augmentation for multi-modal data is still an open question in the research community. We thus do not consider data augmentation in the baseline and will leave it in our future work. 

Finally, domain adaptation baselines comprise \textbf{DAN} \cite{DAN}, \textbf{DANN} \cite{DANN}, \textbf{Coral} \cite{coral} and $\mathbf{M}^3\mathbf{DA}$ \cite{msa1cvpr19}. \textbf{DAN} and \textbf{DANN} reduce domain discrepancy between the source and target domains by minimizing MMD metric and adversarial learning correspondingly. \textbf{Coral} aligns the second-order statistics of the source and target distributions using a nonlinear transformation. $\mathbf{M}^3\mathbf{DA}$ employs moment matching to align each pair of source domains and each source domain with the target domain. Moreover, it further aligns the conditional probability distribution of output given input. 
\textbf{DAN}, \textbf{DANN}, \textbf{Coral} are single-source DA (SDA) methods, while the other belongs to multi-source DA (MDA) methods.

\subsubsection{Implementation Details}
\begin{enumerate}
    \item \textbf{Model Setting}. We adopt TextCNN and ResNet50 as the backbone framework to extract text and image features and map the features into $d$ dimensions, using corresponding two-layer MLPs, for all models except \textbf{Bert}. Moreover, $d$ is set to 256. TextCNN has three 1D convolutional layers with kernel sizes 3, 4, and 5, and the filter size of each layer is 100. While we finetune ResNet50 for the baseline \textbf{ResNet}, we freeze the weights of this visual encoder for the other models. We initialize TextCNN word embedding in the same way as \textbf{TextCNN-roberta}. As existing domain generalization and domain adaptation methods are devised for only one input modality, we apply these algorithms to the combined features. We concatenate text and image features and then use an external MLP to map them to $d$ dimension. 
    
    \item \textbf{Domain Setting}. We select three events as source domains and the remaining one as the target domain. We combine three source domains as a source domain for SDA baselines (i.e., \textbf{DAN}, \textbf{DANN}, and \textbf{Coral}) while keeping these source domains individual for MDA approaches (i.e., $\mathbf{M}^3\mathbf{DA}$ and our proposed \textbf{RDCM}).
    
    \item \textbf{Training Setting}. The sample size of each domain is set to 32 for each minibatch. For data preprocessing, we first resize the image to $224 \times 224$ and then normalize pixel values to have a mean of [0.485, 0.456, 0.406] and a standard deviation of [0.229, 0.224, 0.225] to ensure compatibility with our visual backbone, ResNet50 \cite{resnet}. For hyperparameters, we fix the sigma of Gaussian kernels as [2, 4, 8, 16] for both modalities (We adopt multi-kernel MMD in our experiments). If not otherwise stated, we set the threshold $\beta$ in Eq.~\ref{indicator} to 0.5 and the temperature $\tau$ in Eq.~\ref{intra} to 0.5. Moreover, we only finetune the weights of different losses $\lambda_1$ and $\lambda_2$ for our model by searching from [0.005, 0.1, 0.5, 1, 5, 10]. At last, We find that $\lambda_1=0.1$ and $\lambda_2=0.5$ achieve the best performance on the Pheme dataset, while $\lambda_1=1$ and $\lambda_2=1$ are optimal for the Twitter dataset and Cross dataset. We mainly finetune the loss weights for baselines by searching from [0.01, 0.1, 1, 10, 100, 1000] to find the best hyperparameter. We adopt Adam as the optimizer with a learning rate 0.001 and weight decay of 0.0005. All models are trained for 20 epochs on the Pheme and Cross datasets and 30 epochs on the Twitter dataset. 

    \item \textbf{Evaluation Protocol}. We utilize accuracy as the evaluation metric. In our work, we follow existing work in the community of domain generalization and domain adaptation \cite{dghaoliang1, motiian2017unified} and use the standard evaluation protocol. Especially, for each dataset, we divide each domain into a training set (70\%) and a test set (30\%) via random selection from the overall dataset and conduct a leave-one-domain-out evaluation. In domain generalization, we use the training split of source domains to train and select the optimal model based on the validation results of the testing split of source domains, while we employ the training split of source samples and the unlabelled target domain examples to train and also validate the model on the testing split in domain adaptation. For testing, we evaluate the model on the entire target domain for DG and DA. To avoid randomness, all experiments are repeated three times with different random seeds, and the average result and standard deviation are reported.
\end{enumerate}

\begin{table*}
  \caption{Pheme dataset results of four groups of approaches.}
  \centering
  \label{main_results_1}
  \begin{tabular}{lc|cccc|c}
    \hline
    \multicolumn{2}{c}{Model}&$\mathcal{COF}\rightarrow \mathcal{S}$  (\%)&$\mathcal{CSF}\rightarrow \mathcal{O}$(\%)&$\mathcal{CSO}\rightarrow \mathcal{F}$	(\%)&$\mathcal{OFS}\rightarrow \mathcal{C}$(\%)&Avg(\%)\\
    \hline
    \multirow{4}*{\makecell[c]{Uni-modality}}
    &TextCNN-rand&56.41$\pm$1.9&52.43$\pm$4.2&86.74 $\pm$2.4&79.33$\pm$1.5&68.72$\pm$2.2\\
    &TextCNN-roberta&\textbf{62.38}$\pm$1.4&\textbf{64.24}$\pm$1.0&\textbf{87.95}$\pm$0.2&\textbf{81.95}$\pm$0.3&\textbf{74.13}$\pm$0.3\\
    &Bert\cite{bert}&60.53$\pm$2.4&57.29$\pm$1.0&79.72$\pm$2.6&78.72$\pm$0.4&69.07$\pm$1.5\\
    &ResNet\cite{resnet}&56.22$\pm$0.7&47.18$\pm$2.0&86.45$\pm$2.9&70.90$\pm$3.2&65.19$\pm$1.7\\
    \hline
    \multirow{2}*{\makecell[c]{Multi-modality}}
    &Vanilla \cite{vanilla}&\textbf{65.79}$\pm$1.7&\textbf{64.67}$\pm$2.2&87.45$\pm$0.2&\textbf{81.02}$\pm$0.5&\textbf{74.73}$\pm$0.8\\
    &ModalityGat \cite{multimodalgat}&56.09$\pm$2.3&47.48$\pm$5.0&\textbf{88.03}$\pm$0.0&80.32$\pm$0.2&67.98$\pm$1.3\\
    \hline
    \multirow{5}*{\makecell[c]{Domain Generalization}
    }
    &EANN  \cite{drkdd2018eann} &65.97$\pm$1.3&65.62$\pm$2.9&88.07$\pm$0.5&80.42$\pm$0.0&75.02$\pm$0.9\\
    &IRM \cite{irm}&65.02$\pm$0.6&64.71$\pm$1.7&87.50$\pm$1.0&81.23$\pm$0.2&74.64$\pm$0.2\\
    &MLDG \cite{mldg}&64.41$\pm$2.4&64.84$\pm$0.4&88.35$\pm$0.2&81.56$\pm$0.1&74.79$\pm$0.5\\
    &Fish \cite{fish} &55.87$\pm$2.1&43.58$\pm$0.5&88.03$\pm$0.0&75.88$\pm$4.7&65.84$\pm$0.6\\
    \cdashline{2-7}[1pt/1pt]
    &\textbf{RDCM}(DG)&\textbf{67.36}$\pm$1.8&\textbf{66.49}$\pm$2.7&\textbf{88.41}$\pm$0.6&\textbf{81.89}$\pm$0.0&\textbf{76.04}$\pm$0.9\\
    \hline
    \multirow{5}*{\makecell[c]{Domain Adaptation}
    }
    &DAN \cite{DAN}&67.09$\pm$0.3&62.46$\pm$1.4&86.04$\pm$2.4&80.56$\pm$0.2&74.04$\pm$0.9\\
    &DANN \cite{DANN}&69.24$\pm$1.2&64.67$\pm$2.8&87.66$\pm$0.6&81.29$\pm$0.2&75.72$\pm$1.1\\
    &Coral \cite{coral}&\textbf{69.66}$\pm$0.7&64.19$\pm$2.4&85.60$\pm$2.5&80.70$\pm$0.0&75.04$\pm$1.0\\
    &$\textrm{M}^3\textrm{DA}$ \cite{msa1cvpr19} &66.75$\pm$1.7&66.63$\pm$0.6&88.20$\pm$0.4&81.06$\pm$0.3&75.66$\pm$0.6\\
    \cdashline{2-7}[1pt/1pt]
    &\textbf{RDCM}(DA)&67.49$\pm$1.7&\textbf{68.75}$\pm$1.0&\textbf{88.48}$\pm$0.0&\textbf{82.16}$\pm$0.3&\textbf{76.72}$\pm$0.3\\
  \hline
\end{tabular}
\end{table*}

\begin{table*}
   \caption{Twitter dataset results of four groups of approaches.}
  \centering
  \label{main_results_2}
  \begin{tabular}{lc|cccc|c}
    \hline
    \multicolumn{2}{c}{Model}&$\mathcal{ABI}\rightarrow \mathcal{M}$  (\%)&$\mathcal{BMI}\rightarrow \mathcal{A}$(\%)&$\mathcal{AMI}\rightarrow \mathcal{B}$	(\%)&$\mathcal{ABM}\rightarrow \mathcal{I}$(\%)&Avg(\%)\\
    \hline
    \multirow{4}*{\makecell[c]{Uni-modality}}
    &TextCNN-rand&45.95$\pm$3.0&53.12$\pm$1.2&59.82$\pm$4.6&46.80$\pm$2.5&51.42$\pm$0.9\\
    &TextCNN-roberta&46.31$\pm$0.7&56.12$\pm$0.4&69.76$\pm$1.1&40.01$\pm$1.0&53.05$\pm$0.5\\
    &Bert\cite{bert}&58.44$\pm$2.9&54.55$\pm$0.7&75.27$\pm$2.1&\textbf{55.51}$\pm$3.6&60.94$\pm$1.3\\
    &ResNet\cite{resnet}&\textbf{76.89}$\pm$4.6&\textbf{54.73}$\pm$3.1&\textbf{83.40}$\pm$0.3&36.71$\pm$2.6&\textbf{62.93}$\pm$2.1\\
    \hline
    \multirow{2}*{\makecell[c]{Multi-modality}}
    &Vanilla \cite{vanilla}&81.44$\pm$1.0&\textbf{61.11}$\pm$4.8&79.31$\pm$1.5&\textbf{40.12}$\pm$2.3&\textbf{65.50}$\pm$1.1\\
    &ModalityGat \cite{multimodalgat}&\textbf{86.32}$\pm$0.3&59.55$\pm$0.3&\textbf{80.62}$\pm$0.2&34.61$\pm$3.1&65.28$\pm$0.6\\
    \hline
    \multirow{5}*{\makecell[c]{Domain Generalization}}
    &EANN  \cite{drkdd2018eann} &88.42$\pm$3.5&56.61$\pm$0.2&71.57$\pm$4.3&57.25$\pm$2.9&68.46$\pm$1.9\\
    &IRM \cite{irm}&71.88$\pm$2.7&53.13$\pm$0.2&80.24$\pm$0.3&\textbf{58.36}$\pm$0.4&65.90$\pm$1.0\\
    &MLDG \cite{mldg}&86.25$\pm$6.5&56.23$\pm$0.7&78.94$\pm$0.2&51.20$\pm$8.7&68.16$\pm$3.6\\
    &Fish \cite{fish} &71.86$\pm$5.3&55.61$\pm$0.0&79.56$\pm$0.5&45.11$\pm$6.6&63.03$\pm$3.8\\
    \cdashline{2-7}[1pt/1pt]
    &\textbf{RDCM}(DG)&\textbf{88.49}$\pm$0.7&\textbf{58.15}$\pm$1.9&\textbf{81.32}$\pm$1.8&52.48$\pm$2.3&\textbf{70.11}$\pm$0.6\\
    \hline
    \multirow{5}*{\makecell[c]{Domain Adaptation}}
    &DAN \cite{DAN}&89.37$\pm$1.0&58.29$\pm$0.7&77.80$\pm$1.6&44.21$\pm$4.7&67.42$\pm$2.5\\
    &DANN \cite{DANN}&89.49$\pm$1.0&60.01$\pm$0.2&78.27$\pm$1.8&49.62$\pm$3.3&69.35$\pm$2.1\\
    &Coral \cite{coral}&89.91$\pm$0.3&60.38$\pm$1.7&78.41$\pm$1.5&47.52$\pm$5.8&69.05$\pm$2.8\\
    &$\textrm{M}^3\textrm{DA}$ \cite{msa1cvpr19} &89.99$\pm$3.2&55.94$\pm$0.7&79.35$\pm$0.8&\textbf{55.53}$\pm$2.0&70.20$\pm$1.3\\
    \cdashline{2-7}[1pt/1pt]
    &\textbf{RDCM}(DA)&\textbf{90.11}$\pm$0.6&\textbf{60.78}$\pm$1.4&\textbf{79.47}$\pm$1.9&55.50$\pm$3.1&\textbf{71.47}$\pm$0.7\\
  \hline
\end{tabular}
\end{table*}

\begin{table*}
  \caption{Cross-dataset results of four groups of approaches.}
  \centering
  \label{crossset}
  \begin{tabular}{lc|cccc|c}
    \hline
    \multicolumn{2}{c}{Model}&$\mathcal{COF}\rightarrow \mathcal{M}$  (\%)&$\mathcal{CSF}\rightarrow \mathcal{A}$(\%)&$\mathcal{ABI}\rightarrow \mathcal{S}$	(\%)&$\mathcal{ABI}\rightarrow \mathcal{O}$(\%)&Avg(\%)\\
    \hline
    \multirow{2}*{\makecell[c]{Uni-modality}}
    &TextCNN-roberta&49.74$\pm$0.4&56.11$\pm$0.1&53.84$\pm$1.3&\textbf{52.28}$\pm$1.2&52.99$\pm$1.6\\
    &ResNet\cite{resnet}&\textbf{53.67}$\pm$0.4&\textbf{58.32}$\pm$0.9&\textbf{58.02}$\pm$1.1&49.87$\pm$0.3&\textbf{54.97}$\pm$0.3\\
    \hline
    \multirow{2}*{\makecell[c]{Multi-modality}}
    &Vanilla~\cite{vanilla}&\textbf{48.66}$\pm$2.8&\textbf{57.28}$\pm$0.3&\textbf{59.40}$\pm$1.0&48.52$\pm$1.5&\textbf{53.47}$\pm$1.3\\
    &ModalityGat~\cite{multimodalgat}&38.46$\pm$0.5&55.86$\pm$0.2&56.14$\pm$0.4&\textbf{52.08}$\pm$1.8&50.64$\pm$0.6\\

    \hline
    \multirow{5}*{\makecell[c]{Domain Generalization}
    }
    &EANN  \cite{drkdd2018eann} &52.23$\pm$4.4&57.01$\pm$0.2&58.34$\pm$1.6&52.98$\pm$2.8&55.14$\pm$1.9\\
    &IRM \cite{irm}&52.93$\pm$3.7&56.11$\pm$0.6&57.16$\pm$0.9&53.03$\pm$0.0&54.81$\pm$1.9\\
    &MLDG \cite{mldg}&53.30$\pm$0.6&55.28$\pm$0.1&56.64$\pm$1.0&52.82$\pm$0.6&54.51$\pm$0.5\\
    &Fish \cite{fish} &47.78$\pm$1.5&51.23$\pm$4.6&53.49$\pm$2.3&48.00$\pm$3.4&50.12$\pm$2.6\\
    \cdashline{2-7}[1pt/1pt]
    &\textbf{RDCM}(DG)&\textbf{53.41}$\pm$1.7&\textbf{57.40}$\pm$0.2&\textbf{59.90}$\pm$2.0&\textbf{53.17}$\pm$0.5&\textbf{55.97}$\pm$1.2\\
    \hline
    \multirow{5}*{\makecell[c]{Domain Adaptation}
    }
    &DAN \cite{DAN}&53.29$\pm$0.8&57.26$\pm$0.2&59.12$\pm$2.0&51.57$\pm$1.8&55.31$\pm$0.5\\
    &DANN \cite{DANN}&54.66$\pm$6.7&57.03$\pm$0.8&55.10$\pm$0.5&51.08$\pm$1.3&54.47$\pm$2.2\\
    &Coral \cite{coral}&54.20$\pm$3.1&58.01$\pm$1.0&56.36$\pm$1.5&51.48$\pm$2.4&55.01$\pm$1.2\\
    &$\textrm{M}^3\textrm{DA}$ \cite{msa1cvpr19} &53.61$\pm$1.8&58.36$\pm$0.3&58.84$\pm$1.0&51.34$\pm$1.2&55.54$\pm$1.0\\
    \cdashline{2-7}[1pt/1pt]
    &\textbf{RDCM}(DA)&\textbf{55.27}$\pm$2.6&\textbf{58.49}$\pm$0.5&\textbf{60.33}$\pm$0.6&\textbf{52.00}$\pm$2.5&\textbf{56.52}$\pm$1.0\\
  \hline
\end{tabular}
\end{table*}

\begin{table}
  \caption{$\mathcal{A}$-distance of four cases for Pheme and Twitter datasets in DG and DA settings.}
  \centering
  \label{adistance}
  \begin{tabular}{cccccc}
    \hline
        \multicolumn{6}{c}{Pheme Dataset}\\
      \hline
    Model&Metric&$\mathcal{S}$  &$\mathcal{O}$&$\mathcal{F}$	&$\mathcal{C}$\\
      \cline{1-6}
    \multirow{2}*{\makecell[c]{\textbf{RDCM}(DG)}}&Acc(\%)&67.36&66.49&88.41&81.89\\
    &$\mathcal{A}$-distance&1.79&1.78&1.73&1.76\\
    \cline{1-6}
    \multirow{2}*{\makecell[c]{\textbf{RDCM}(DA)}}&Acc(\%)&67.49&68.75&88.48&82.16\\
    &$\mathcal{A}$-distance&1.75&1.76&1.64&1.73\\
    \hline
    \multicolumn{6}{c}{Twitter Dataset}\\
      \hline
    Model&Metric&$\mathcal{M}$  &$\mathcal{A}$&$\mathcal{B}$	&$\mathcal{I}$\\
     \cline{1-6}
    \multirow{2}*{\makecell[c]{\textbf{RDCM}(DG)}}&Acc(\%)&88.49&58.15&81.32&52.48\\
    &$\mathcal{A}$-distance&1.69&1.62&1.64&1.90\\
     \cline{1-6}
    \multirow{2}*{\makecell[c]{\textbf{RDCM}(DA)}}&Acc(\%)&90.11&60.78&79.47&55.50\\
    &$\mathcal{A}$-distance&1.64&1.68&1.61&1.89\\
  \hline
\end{tabular}
\end{table}

\subsection{RQ1: Effectiveness of data collected from unlabeled target domain and multiple modalities}
There are two motivations for our work. First, existing robust domain misinformation detection methods do not consider the dynamic propagation trend of online information. In other words, it is necessary to cover DG and DA for our method based on the availability of the target domain. Accordingly, an indispensable premise is that the target domain data could further boost the detection performance compared with DG. On the other hand, fewer recent approaches concentrate on the importance of the semantic gap between textual and visual modalities. However, a foundation of this motivation is that multi-modal methods could have advantages over uni-modal ones. As a result, we conduct comprehensive experiments and report the accuracy and standard error in Table \ref{main_results_1} and Table \ref{main_results_2}, aiming to prove the validity of both motivations.

\subsubsection{Importance of the Target Domain}

We show the impact of unlabeled target domain data for improving the performance of misinformation detection. Some theoretical analyses~\cite{ben2010theory, msatheory1, msatheory2} bound the target error in terms of the source error, the divergence between the distributions of the source domain and the target domain, and other components. In other words, when reducing the discrepancy among source domains, we could improve the classification accuracy in the target domain by concurrently reducing the discrepancy between the target domain and source domains. In turn, we conduct two-sided Wilcoxon rank-sum statistic\footnote{\url{https://data.library.virginia.edu/the-wilcoxon-rank-sum-test/}. The Wilcoxon Rank Sum Test is the non-parametric version of the two-sample t-test, which works when our samples are small.} for the average accuracy of DG and DA baselines on two datasets. The p-values of our tests (0.25 for the Pheme dataset and 0.12 for the Twitter Dataset) are more than 0.05. 

\subsubsection{Effectiveness of Multi-modal Methods}
We illustrate the superiority of exploiting both modalities by analyzing the experimental results of unimodal and multi-modal methods. On the Pheme dataset, \textbf{Vanilla}, combining textual and visual features surpasses \textbf{TextCNN-roberta} with 0.60\% improvement and \textbf{Resnet} with 9.54\%. When on the Twitter dataset, this multi-modal method also brings 2.57\% improvement compared with \textbf{ResNet}.

\textbf{ResNet} shows the opposite trend. It is possibly due to differences between two datasets, such as data collection ways and label protocols, which is a common case for practical applications. Especially, advisable multi-modal models could have the potential to combine complementary information from multiple modalities by filtering noise and resolving conflicts based on comprehending correlations between these modalities, which justifies the advantage of exploiting both texts and images for our task. We adopt \textbf{Vanilla} as the backbone for subsequent experiments. 

\begin{shaded*}
{\noindent \textbf{Answer to RQ1}: Target domain and multi-modal inputs effectively aid robust domain misinformation detection.}
\end{shaded*}

\subsection{RQ2: Effectiveness of Our Method}
Given the news propagation dynamics, it would be beneficial for robust domain approaches to cover domain adaptation and domain generalization simultaneously. To verify the effectiveness and versatility of our method for both settings, we compare \textbf{RDCM} with \textbf{Vanilla}, DG baselines, and DA baselines. Table~\ref{main_results_1}, Table~\ref{main_results_2} and Table~\ref{crossset} show such results. 

We first discuss the comparisons with \textbf{Vanilla}. On the Pheme dataset, the DG and DA versions of \textbf{RDCM} outperform \textbf{Vanilla} by 1\%. And the superiority is more significant for the Twitter and Cross datasets. It evinces that inter-domain alignment and cross-modality alignment modules positively influence discriminating the misinformation. 

Regarding DG baselines, \textbf{RDCM} consistently outperforms most of them by a clear margin and simultaneously achieves over 1\% improvement compared with SOTA \textbf{EANN} on Pheme and Twitter datasets. Similarly, our proposed method also outperforms all DA baselines on three datasets.  We suggest two possible reasons. First, we employ the kernel mean embedding to represent the joint distribution of textual and visual variables to perform domain alignment, which can capture the correlation between variables \cite{mmd1,dghaoliang1} to reduce incorrect classification. Second, we further mitigate the semantic gap between text and image modalities based on contrastive learning to enable cross-modal misinformation detection compared to other baselines. We also observe that multi-source DA methods (e.g., $\mathbf{M}^3\mathbf{DA}$ and \textbf{RDCM}) perform better than single-source DA methods (e.g., \textbf{DAN}, \textbf{DANN}, and \textbf{Coral}). Hence, we devise our inter-domain alignment component in the multi-source DA version.

Additionally, it is worth noting that the performance of the proposed method and the baselines significantly differ among the four target domains. For instance, we observe that our model performs better on cases $\mathcal{CSO}\rightarrow \mathcal{F}$ and $\mathcal{OFS}\rightarrow \mathcal{C}$
than it does on $\mathcal{COF}\rightarrow \mathcal{S}$ and $\mathcal{CSF}\rightarrow \mathcal{O}$. We suggest two possible causes for this phenomenon: 1) The domain gap between source and target domains for cases with poor generalization performance may be larger than those cases where models can generalize well. That is because the generalization performance largely depends on domain discrepancies between source domains and the target domain \cite{dgthory1, ben2010theory}. To validate this conjecture, we exploit $\mathcal{A}$-distance\footnote{$\mathcal{A}$-distance is defined as $\hat{d}_\mathcal{A}=2(1-\epsilon$) where $\epsilon$ is the generalization error of a two-sample classifier (kernel SVM in our case, following \cite{DAN})  trained on the binary problem to
distinguish input samples between the source and target
domains.}, presented by Ben-David et al.\cite{ben2010theory}, to measure domain discrepancies for different cases of Pheme and Twitter datasets in Tabel~\ref{adistance}. The results show that models can learn more domain-invariant features in component cases than problematic ones to prove our conjectures. 2) In bad cases, the target domains may be more challenging and complex. For instance, the tweets labeled as rumors in $\mathcal{S}$ and $\mathcal{O}$ have more diversified styles and patterns \cite{review24pheme}. As a result, it is difficult for models trained on source domains to learn beneficial invariance capable of covering the distribution of the intractable target domain.

\begin{shaded*}
{\noindent \textbf{Answer to RQ2}: The proposed methods generally outperform different backbone networks, as well as all DG and DA baseline models based on two different settings, which evinces the effectiveness of our proposed \textbf{RDCM}. }
\end{shaded*}

\subsection{RQ3: Analysis of Different Components}
In this subsection, we conduct an ablation study to understand the impact of  Inter-domain and Cross-modality Alignment modules of our proposed method. For brevity, we only report detection accuracy in DG.
\begin{table}
   \caption{Experimental results of ablation study in domain generalization.}
  \centering
  \label{ablation_results}
  \begin{tabular}{c|cccc|c}
    \hline
    \multicolumn{6}{c}{Pheme Dataset}\\
      \hline
     &$ \mathcal{S}$(\%)&$\mathcal{O}$(\%)&$\mathcal{F}$(\%)&$\mathcal{C}$(\%)&Avg(\%)\\
      \hline
    \textbf{Ours}(DG)&\textbf{67.36}&\textbf{66.49}&\textbf{88.41}&81.89&\textbf{76.04}\\
    \cdashline{1-6}[1pt/1pt]
     w/o inter&66.61&66.17&88.30&81.47&75.64\\
     w/o cross&67.33&65.41&88.19&\textbf{81.98}&75.73\\
    w/o both&65.78&64.67&87.45&81.02&74.73\\
    \hline
     \multicolumn{6}{c}{Twitter Dataset}\\
     \hline
    &$ \mathcal{M}$(\%)&$\mathcal{A}$(\%)&$\mathcal{B}$(\%)&$\mathcal{I}$(\%)&Avg(\%)\\
     \hline
     \textbf{Ours}(DG)&88.49&58.15&\textbf{81.32}&\textbf{52.48}&\textbf{70.11}\\
    \cdashline{1-6}[1pt/1pt]
     w/o inter&82.63&56.00&75.23&51.48&66.34\\
     w/o cross&\textbf{91.08}&57.81&80.85&45.10&68.71\\
    w/o both&81.44&\textbf{61.11}&79.31&40.12&65.50\\
  \hline
\end{tabular}
\end{table}

We consider three variants, including removing the inter-domain alignment component (denoted as w/o inter), removing the cross-modality alignment component (denoted as w/o cross), and removing both components (denoted as w/o both). The results in Table~\ref{ablation_results} are telling. Despite the performance drop in certain cases (e.g., $\mathcal{M}$ and $\mathcal{A}$ in the Twitter dataset) compared to other baselines, our model generally performs best when leveraging all these components. It suggests that our model benefits from both alignment modules. Moreover, \textbf{Ours} may overfit to cross-modality alignment loss for $\mathcal{M}$ and overfit to both inter-domain alignment and cross-modality alignment losses for  $\mathcal{A}$, which can be mitigated by adjusting weights of different loss ($\lambda_1$ and  $\lambda_2$). In turn, removing inter-domain alignment leads to a greater performance drop than cross-domain alignment, especially in the Twitter dataset. However, it is difficult to determine which is more important because of the comparable performance on the Pheme dataset. 

\begin{shaded*}
{\noindent \textbf{Answer to RQ3}: Each component of \textbf{RDCM} contributes positively to multi-modal misinformation detection task. Both components are important and could be assisted by each other.}
\end{shaded*}
\begin{figure}[t]
\centering
\begin{subfigure}[b]{0.235\textwidth}
\centering
\includegraphics[width=\linewidth]{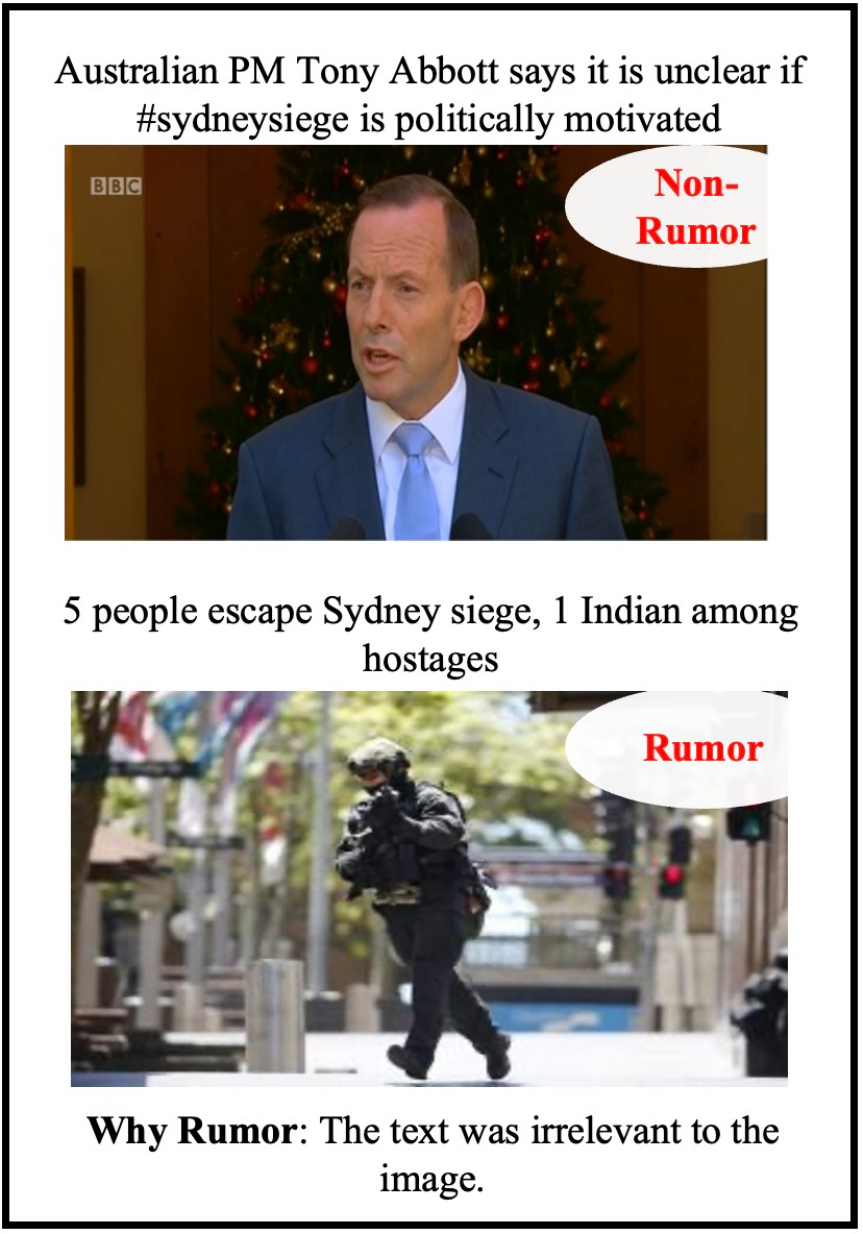}
\caption{$\mathcal{COF}\rightarrow \mathcal{S}$}
\label{sspheme}
\end{subfigure}
\hfill
\begin{subfigure}[b]{0.235\textwidth}
\includegraphics[width=\linewidth]{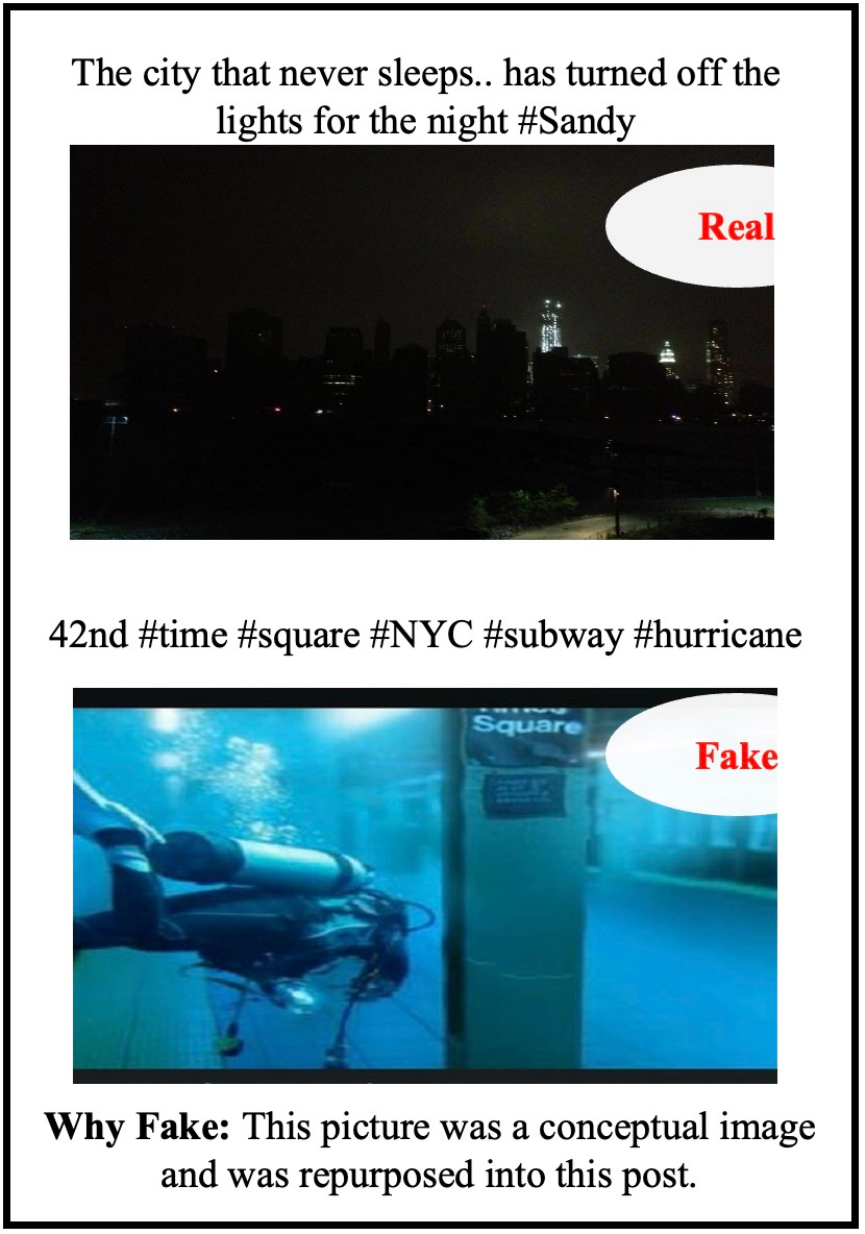} 
\centering
\caption{$\mathcal{BMI}\rightarrow \mathcal{A}$}
\label{sstwitter}
\end{subfigure}
\caption{The examples of Sydney Siege from $\mathcal{COF}\rightarrow \mathcal{S}$ in the Pheme Dataset and Hurricane Sandy from $\mathcal{BMI}\rightarrow \mathcal{A}$ in the Twitter Dataset. These samples are wrongly classified by \textbf{Vanilla} but can be identified correctly by our proposed \textbf{RDCM} (DG). Sydney Siege was a terrorist attack that a gunman held hostage ten customers and eight employees in Sydney on December 15-16, 2014. Moreover, Hurricane Sandy was extremely destructive and strong, affecting 24 states in the United States.
}
\label{casestudy}
\end{figure}
\subsection{Case Study}
To further justify the effectiveness of our proposed model RDCM (DG), we provide case studies on samples that are misclassified by \textbf{Vanilla} \cite{vanilla} but are detected accurately by our proposed model, which incorporates domain alignment and cross-modal alignment modules.

As depicted in Fig.~\ref{casestudy}, \textbf{RDCM} excels at understanding semantic correspondences and contradictions between texts and images and learns more transferable implicit patterns for multimodal misinformation detection compared to \textbf{Vanilla}. For instance, identifying non-rumor and real samples may imply that our model can comprehend that ``Australian PM Tony Abbott" and ``turn off the lights"  align with the person and the dark background in the attached images, respectively.  Additionally, identifying the rumor sample in Fig.~\ref{sspheme} depends on spotting the cross-modal irrelevance. We suggest the success of these samples may stem from our cross-modal alignment component that effectively reduces the modality gap through contrastive learning. Moreover, our model may learn contributive domain-invariant features better, such as the races of artificial synthesis as shown in the fake sample in Fig.~\ref{sstwitter}, owing to the inter-domain alignment module aligning the joint distribution of both modalities conditioned on their correlation information.

\section{Conclusions \& Future work}
In this paper, we tackled the problem of robust domain misinformation detection. We presented a robust domain and modality-alignment framework based on inter-domain and cross-modality alignment modules. 

The kernel mean embedding underpins inter-domain alignment to represent the joint distribution of textual and visual modalities. It reduces the domain shift by minimizing the Maximum Mean Discrepancy between the joint distributions. 

The cross-modality alignment module leverages a specific sample strategy to construct positive and negative samples and mitigate the modality gap based on contrastive learning. Experimental results show the effectiveness of the proposed method for robust domain misinformation detection. 

For future work, extending the framework to handle multiple images and long-paragraph texts represents a key step forward. We also suggest exploring various multi-modality scenarios containing video and audio information to enrich the current text- and image-based representations. 

\vspace{-10pt}
\section{Limitations}
While the proposed approach (\textbf{RDCM}) demonstrates versatility and effectiveness for the multimodal misinformation detection task in both domain generalization and domain adaptation scenarios, it is important to acknowledge two possible limitations. Firstly, \textbf{RDCM} employs Maximum Mean Discrepancy (MMD) as a metric to measure the domain discrepancy upon the joint distribution of textual and visual modalities. Although MMD offers theoretical merits, it does have certain deficiencies such as the sensitivity to kernel choices and computationally expensive calculations for large high-dimensional datasets (i.e., the computational complexity is O($n^2$) where $n$ represents the sample size) \cite{mmd2, ben2010theory}. Despite these drawbacks, our proposed method outperforms existing approaches in two publicly available datasets when the sigma of Gaussian kernels is fixed for both modalities and each domain contains a limited number of samples, because of the synergy of inter-domain alignment and intra-domain alignment modules.  Secondly, our method specifically focuses on debunking fake image-text pairs.  Nevertheless, the intricate nature of multimodal inputs permitted by social media platforms, such as short videos and emojis, further harms the deployment of our method in the real world. Therefore, we intend to address these two limitations in our future endeavors.

\section{Appendix}
\begin{figure}[t]
\centering
\begin{subfigure}{0.235\textwidth}
\centering
\includegraphics[width=\linewidth]{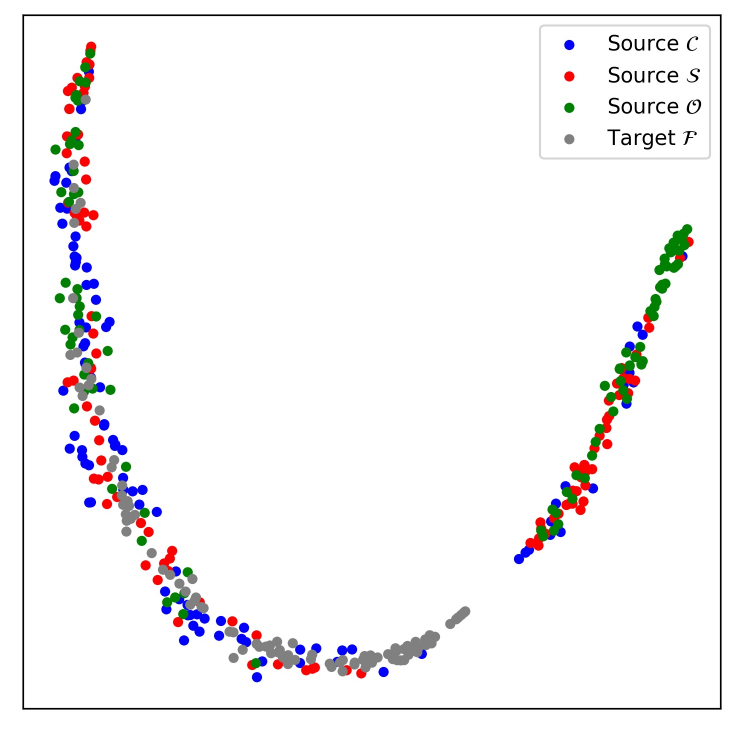}
\caption{\textbf{Joint}}
\label{stj}
\end{subfigure}
\hfill
\begin{subfigure}{0.235\textwidth}
\includegraphics[width=\linewidth]{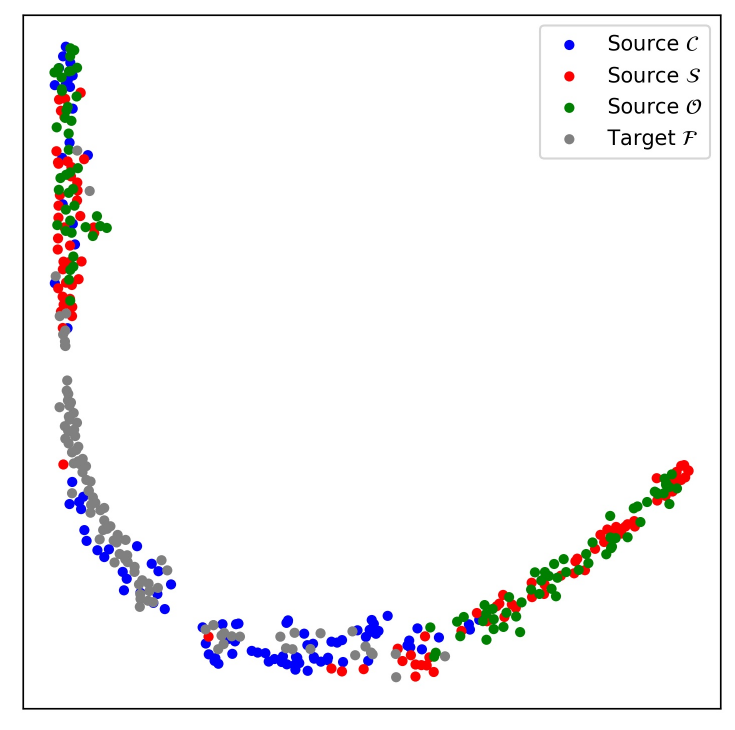} 
\centering
\caption{\textbf{Text}}
\label{stt}
\end{subfigure}
\caption{t-SNE visualization of combined features belonging to three source domains ($\mathcal{C}$, $\mathcal{S}$ and $\mathcal{O}$) and one target domain ($\mathcal{F}$) for Pheme dataset. The features of domains $\mathcal{C}$ and $\mathcal{F}$  are mainly distributed in the two clusters on the left bottom of Fig.~\ref{stt} while the features of these two domains scatter more evenly in Fig.~\ref{stj}.}
\label{tsnsvisulization}
\end{figure}
\subsection{RQ4: Analysis of Inter-domain Alignment}
\begin{table}
   \caption{Comparison results of inter-domain alignment on different modalities in domain generalization.}
  \centering
  \label{mmd_results}
  \begin{tabular}{c|cccc|c}
    \hline
    \multicolumn{6}{c}{Pheme Dataset}\\
     \hline
     Model&$ \mathcal{S}$(\%)&$\mathcal{O}$(\%)&$\mathcal{F}$(\%)&$\mathcal{C}$(\%)&Avg(\%)\\
      \hline
     Fusion&66.72&65.28&88.07&82.10&75.54\\
     Vision&66.64&65.36&88.00&\textbf{82.18}&75.55\\
    Text&64.89&64.84&88.00&81.94&74.92\\
      \cdashline{1-6}[1pt/1pt]
    \textbf{Joint}&\textbf{67.33}&\textbf{65.41}&\textbf{88.19}&81.98&\textbf{75.73}\\
    \hline
     \multicolumn{6}{c}{Twitter Dataset}\\
     \hline
    Model&$ \mathcal{M}$(\%)&$\mathcal{A}$(\%)&$\mathcal{B}$(\%)&$\mathcal{I}$(\%)&Avg(\%)\\
     \hline
     Fusion&89.35&56.62&77.23&45.03&67.06\\ 
     Vision&78.78&56.19&73.18&\textbf{49.85}&64.50\\
    Text&85.98&\textbf{60.07}&80.41&49.15&\textbf{68.90}\\
     \cdashline{1-6}[1pt/1pt]
     \textbf{Joint}&\textbf{91.08}&57.81&\textbf{81.25}&44.70&68.71\\
  \hline
\end{tabular}
\end{table}
In Inter-domain Alignment, we assume the domain shift exists in the joint distribution of multiple modalities instead of the marginal distribution of any individual modality. Furthermore, unlike simple fusion (e.g., concatenation), we employ the kernel mean embedding to represent the joint distribution. To show the superiority of this module, we conduct experiments on four models. The first model, \textbf{Fusion}, involves aligning the joint distribution of both modalities obtained by concatenation, described as $\textrm{MMD}(\mathcal{D}_S^i,\mathcal{D}_S^j) = \|\mu_{\mathbf{X}_{t,v,i}}-\mu_{\mathbf{X}_{t,v,j}}\|_\mathcal{H}^2$ where $\mathbf{X}_{t,v}$ represents the random variable of the concatenation of textual and visual features. The second, \textbf{Vision}, aligns the marginal distribution upon visual features, described as $\small \textrm{MMD}(\mathcal{D}_S^i,\mathcal{D}_S^j) =  \|\mu_{\mathbf{X}_{v,i}}-\mu_{\mathbf{X}_{v,j}}\|_\mathcal{H}^2$. The third one, \textbf{Text}, aligns marginal distribution upon textual features, described as $ \small \textrm{MMD}(\mathcal{D}_S^i,\mathcal{D}_S^j) = \|\mu_{\mathbf{X}_{t,i}}-\mu_{\mathbf{X}_{t,j}}\|_\mathcal{H}^2$. Finally, the fourth one,  \textbf{Joint}, aligns the joint distribution of both modalities obtained by our proposed kernel mean embedding in Eq. \ref{jkernelemb}, described as $\textrm{MMD}(\mathcal{D}_S^i,\mathcal{D}_S^j) = \|\mu_{\mathbf{X}_{t,i},\mathbf{X}_{v,i}}-\mu_{\mathbf{X}_{t,j},\mathbf{X}_{v,j}}\|_\mathcal{H}^2$. 

From Table~\ref{mmd_results}, we observe that \textbf{Joint} and \textbf{Fusion} usually have higher accuracy than \textbf{Text} and \textbf{Image}, which illustrates the effectiveness of aligning the joint distribution. It may be because deciding which modality mainly accommodates the domain shift is impractical. We further visualize the combined features of different domains extracted by \textbf{Joint} and \textbf{Text} using t-SNE embeddings in Fig. \ref{stj} and Fig. \ref{stt}, respectively. The figures show that the features are less discriminative when generated by \textbf{Joint}, especially for features of the target domain. It also suggests that the adaptation of joint distributions is more powerful than marginal distributions for our task. Besides, the boost of \textbf{Joint} is more significant than \textbf{Fusion}. Such empirical results and theoretical guarantees in Eq.~\ref{jkernelemb} imply that the kernel mean embedding is more effective in modeling the joint distribution for our task.

\begin{shaded*}
{\noindent \textbf{Answer to RQ4}: Aligning the joint distribution of textual and visual modalities achieves better performance than aligning their marginal distributions. Moreover, the mean kernel embedding is more advantageous for modeling the joint distribution compared with fusion through feature concatenation .}
\end{shaded*}

\subsection{RQ5: Analysis of Cross-modality Alignment}
\begin{table}
   \caption{Comparison results of different contrastive learning methods in domain generalization.}
  \centering
  \label{ablation_con}
  \begin{tabular}{c|cccc|c}
    \hline
    \multicolumn{6}{c}{Pheme Dataset}\\
     \hline
    Model&$ \mathcal{S}$(\%)&$\mathcal{O}$(\%)&$\mathcal{F}$(\%)&$\mathcal{C}$(\%)&Avg(\%)\\
      \hline
     Regular \cite{conbaseline1}&58.74&45.23&88.03&80.37&68.09\\
    TextCon&65.66&65.36&88.06&81.47&75.21\\
    ThresCon&64.76&65.41&88.00&80.63&74.70\\
      \cdashline{1-6}[1pt/1pt]
    \textbf{Ours}&\textbf{66.61}&\textbf{66.17}&\textbf{88.30}&\textbf{81.47}&\textbf{75.64}\\
    \hline
     \multicolumn{6}{c}{Twitter Dataset}\\
     \hline
    Model&$ \mathcal{M}$(\%)&$\mathcal{A}$(\%)&$\mathcal{B}$(\%)&$\mathcal{I}$(\%)&Avg(\%)\\
     \hline
    \cdashline{1-6}[1pt/1pt]
     Regular \cite{conbaseline1}&56.78&\textbf{60.42}&70.74&44.53&58.12\\
     TextCon&77.18&56.21&73.48&50.90&64.44\\
    ThresCon&73.76&56.03&70.85&49.57&62.55\\
     \cdashline{1-6}[1pt/1pt]
    \textbf{Ours}&\textbf{82.63}&56.00&\textbf{75.23}&\textbf{51.48}&\textbf{66.34}\\
  \hline
\end{tabular}
\end{table}

In Cross-modality Alignment, we exclude positive and negative samples of low quality by only taking real posts as positive samples and the negative samples selected by our weighting function in Eq.~\ref{indicator} based on image similarity, respectively. To show the usefulness of this strategy (denoted as \textbf{Ours}),   we compare it with three other kinds of contrastive learning methods. The first one, \textbf{Regular},  uses a common contrastive loss \cite{conbaseline1} based on random sampling. The second, \textbf{TextCon}, includes the weighting function but employs text modality-based similar scores instead. Finally,  \textbf{ThresCon} removes the weighting function term and only considers real posts as positive samples. 

\begin{figure}[t]
\centering
\begin{subfigure}{0.235\textwidth}
\centering
\includegraphics[width=\linewidth]{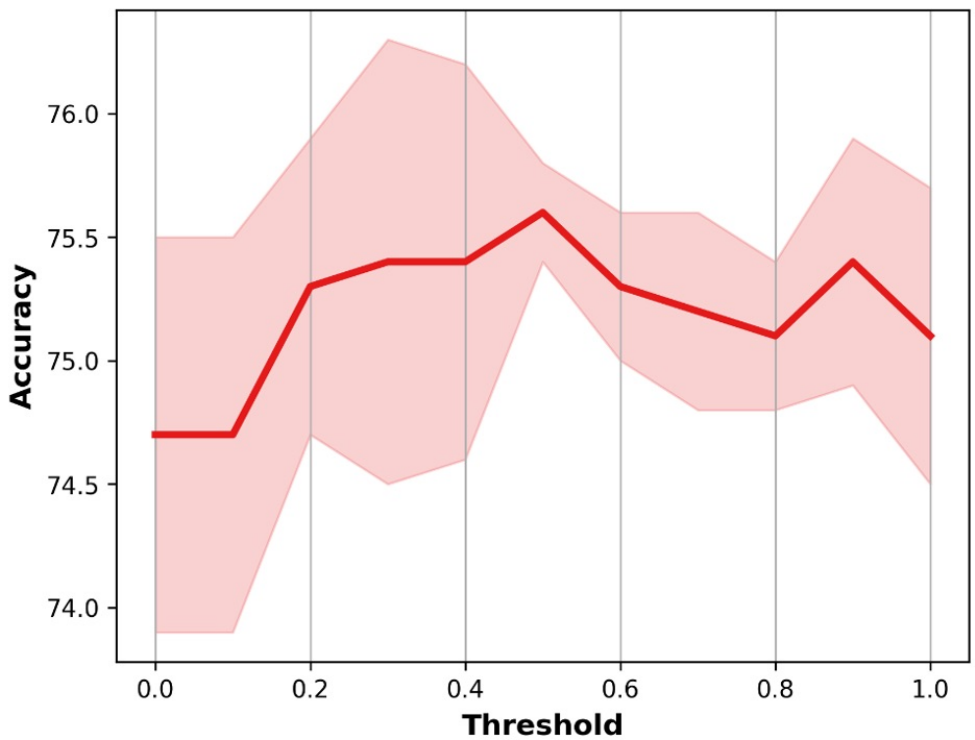}
\caption{Pheme Dataset}
\label{MCA}
\end{subfigure}
\hfill
\begin{subfigure}{0.235\textwidth}
\includegraphics[width=\linewidth]{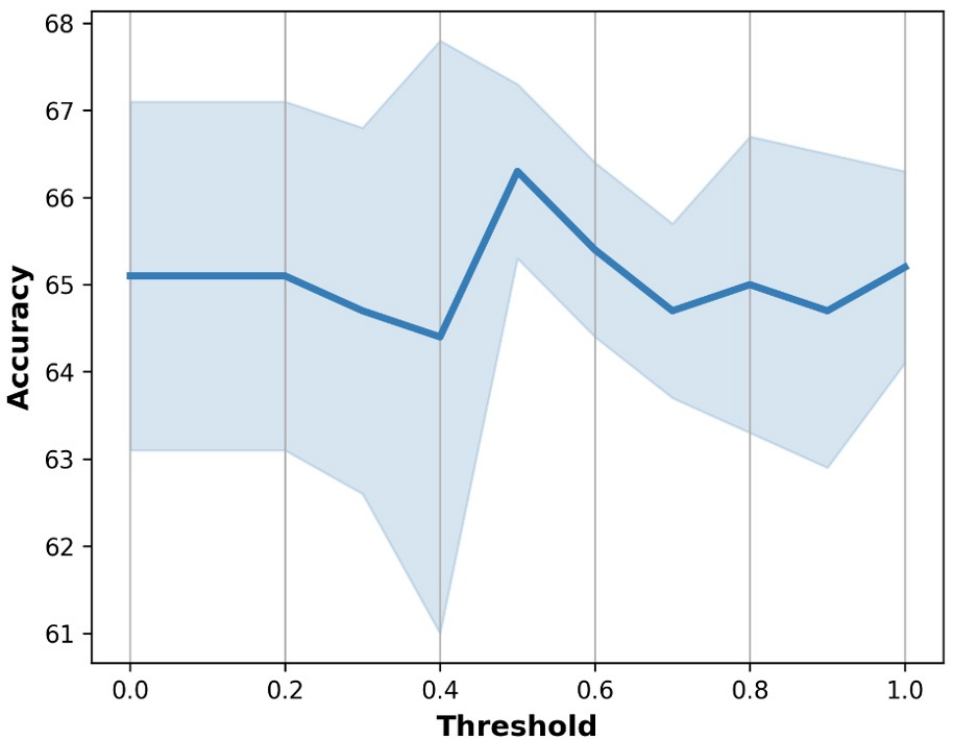} 
\centering
\caption{Twitter Dataset}
\label{gat}
\end{subfigure}
\caption{Performance of our cross-modality alignment module with different thresholds in domain generalization.}
\label{thresfigure}
\end{figure}

As Table~\ref{ablation_con} shows, \textbf{Regular} is dominated by the other three methods by a large margin,  highlighting the importance of filtering out non-relevant samples. Moreover, our method outperforms \textbf{TextCon} and \textbf{ThresCon}, which demonstrates the effectiveness of our proposed indicator function term in Eq. \ref{indicator} that excludes low-quality artificial negative samples based on semantic similarity on the visual modality.
In addition, we conduct experiments with different thresholds (i.e., $\beta$ in Eq.~\ref{indicator}) as Fig.~\ref{thresfigure} depicts. The increase in the threshold brings more noise. This figure shows that the performance first increases and then drops along the threshold increase. Thus, we advocate a tradeoff between sample number and sample noise.

\begin{shaded*}
{\noindent \textbf{Answer to RQ5}: Our model benefits from the proposed sample strategy that can filter non-relevant samples.}
\end{shaded*}

\bibliographystyle{IEEEtran}
\bibliography{IEEEabrv,ref.bib}

\end{document}